\documentclass{article}

\usepackage{PRIMEarxiv}

\usepackage[utf8]{inputenc} 
\usepackage[T1]{fontenc}    
\usepackage{hyperref}       
\usepackage{url}            
\usepackage{booktabs}       
\usepackage{amsfonts}       
\usepackage{nicefrac}       
\usepackage{microtype}      
\usepackage{lipsum}
\usepackage{fancyhdr}       
\usepackage{graphicx}       
\graphicspath{{media/}}     

\usepackage{algorithm}
\usepackage{algorithmic}
\usepackage{amsmath,amssymb,amsfonts}
\usepackage{longtable}
\usepackage{natbib}
\usepackage{makecell}
\usepackage{color,soul}
\usepackage{tabularx, booktabs, makecell}
\usepackage{caption}
\usepackage{subcaption}  
\usepackage{multirow}
\usepackage{enumitem}
\usepackage{listings}
\usepackage[most]{tcolorbox}

\lstdefinestyle{promptstyle}{
  basicstyle=\ttfamily\small,
  breaklines=true,
  columns=fullflexible,
  showstringspaces=false,
  moredelim=**[is][\color{blue!60!black}]{@@}{@@}, 
}

\DeclareCaptionStyle{ruled}{labelfont=normalfont,labelsep=colon,strut=off} 
\floatstyle{ruled}
\newfloat{listing}{tb}{lst}{}
\floatname{listing}{Listing}

\newtcolorbox{promptbox1}[1][]{
  title=Caption Generation Template,
  breakable,
  colback=gray!5,
  colframe=gray!50,
  fonttitle=\bfseries,
  sharp corners,
  enhanced,
  top=6pt,bottom=6pt,left=6pt,right=6pt,
  #1
}

\newtcolorbox{promptbox2}[1][]{
  title=Caption Validation Template,
  breakable,
  colback=gray!5,
  colframe=gray!50,
  fonttitle=\bfseries,
  sharp corners,
  enhanced,
  top=6pt,bottom=6pt,left=6pt,right=6pt,
  #1
}

\usepackage[font=small,labelfont=bf,skip=4pt]{caption}
\usepackage{bbding}

\setlist{noitemsep, topsep=0pt, parsep=0pt, partopsep=0pt, leftmargin=*}

\title{Contrastive Learning with Paraphrasing and Negation}

\author{
  Kwun Ho Ngan\textsuperscript{*}, Joe Townsend \\
  Fujitsu Research of Europe \\
  United Kingdom\\
  \texttt{\{kwun.hongan, joseph.townsend\}@fujitsu.com} \\
   \And
   Saman Sadeghi Afgeh\textsuperscript{*}, Artur d’Avila Garcez \\
  City St George's, University of London \\
  United Kingdom\\
  \texttt{a.garcez@citystgeorges.ac.uk} \\
  \And
  \textsuperscript{*} Equal contribution.
}

\begin{document}
\maketitle

\begin{abstract}
Contrastive vision-language models continue to be the dominant approach for image and text retrieval. Contrastive Language-Image Pre-training (CLIP) trains two neural networks to align their image and text embeddings in a shared latent space. As a challenging case-study for neurosymbolic AI, recent results evaluating CLIP on negated or paraphrased text have shown mixed performance as these are difficult to define formally for text data. Negation produces the opposite meaning using various possible but small lexical changes. Paraphrasing may use very different textual expressions to denote essentially the same thing. As a result, learning of paraphrasing and negation together poses a significant challenge because of the above mismatch between changes in syntax and intended meaning expected to be captured by distances in embedding space. This paper proposes a new CLIP contrastive loss function capable of balancing the requirements of having both paraphrasing and negation. It applies training triplets consisting of original, paraphrased and negated text generated by multiple large language models to the evaluation of CLIP models. The approach, called SemCLIP, aims to learn semantically-relevant and simple embeddings, placing paraphrased captions nearer to the original image embeddings while at the same time pushing negated captions farther away. Empirically, SemCLIP is shown to be capable of preserving roughly the same performance as CLIP augmented with either negation or paraphrasing. Although direct comparisons are difficult to make because the problem of learning with both negation and paraphrasing is different, an expected benefit of SemCLIP should be robustness when applied zero-shot to downstream image classification tasks. Our experiments confirm such robustness as measured by difference in accuracy (mean-accuracy delta) between original and negated captions on five downstream datasets.
\end{abstract}

\section{Introduction}
\label{sec:introduction}

We investigate textual semantic similarity in the context of multimodal representation learning. The dominant approach for image and text learning is based on Contrastive Language-Image Pre-training (CLIP) \citep{radford2021learning}, a modular architecture where two networks processing images and text, respectively, are trained in contrastive manner to align their embeddings into a shared latent space given image-text pairs. Despite CLIP's considerable practical success, it is well-known that negating the meaning of text input, e.g. inserting terms such as ``not'' or ``without'', often fails to produce the desired change in image-text matching. This is not surprising because such minor lexical changes, though semantically significant, produce embedding vectors that remain close by the original text in representation embedding space. This reveals a misalignment during CLIP model training between lexical form and semantic content. Many attempts have been made to refine the learned embeddings to solve this problem \citep{fan2023improving,Patel2024-ur,higgins2018definitiondisentangledrepresentations}. In general, these are either application-specific or make use of pre-defined concepts such as \textit{colour} and \textit{shape} to disentangle embeddings. In this paper, we adopt a different approach inspired by the field of neurosymbolic AI \citep{3rdWave}. We map learned embeddings onto a low-dimensional projection, added naturally as a layer to the network, and use regularization to encourage learning of a clear separation between paraphrases and negation in that low-dimensional space.

Multimodal learning is arguably the most promising frontier for the combined application of learning and reasoning in neural networks, a primary goal of neurosymbolic AI. While language-based reasoning is highly ambiguous and context dependent, making it difficult to provide a formal definition of negation in that context, vision-language models (VLM) can ground textual descriptions into visual context, allowing one modality to help disambiguate the other. CLIP-based learning is especially interesting in terms of neurosymbolic AI due to its modular architecture with separate encoders for text and image. This motivates our exploration of semantic understanding in CLIP under paraphrasing and negation. As a simple example, consider the task of describing the relative positions of objects at a given scene. Assume that the image provides the ground-truth, that is, there is no visual illusion at play. The sentence \textit{a blue square left of a red triangle} can be checked for validity along with paraphrases \textit{a red triangle with a blue square to its left} or \textit{a blue square, with a red triangle to the right of it}. The sentence \textit{a blue square to the right of a red triangle}, however, denotes the opposite, such that if the earlier sentences are \textit{true} then the latter ought to be \textit{false}, assuming no other objects are present in the scene. These examples illustrate the difficulty that CLIP-based models face in detecting small lexical changes that may be semantically critical. It also illustrates the notoriously difficult challenge of formalizing equivalence and inconsistency in natural language, respectively, in the presence of paraphrasing and by defining negation logically. 

To address CLIP's well-known gap between lexical and semantic alignment, CoN-CLIP \citep{singh2024learn} enhanced contrastive learning by incorporating synthetic hard negatives that seek to explicitly encode negation. This approach was motivated by  \citealt{yuksekgonul2023bow}, which showed the difficulty of CLIP at handling negation and applied data augmentation, referred to as NegCLIP, to automatically generate negated counterparts and train CLIP to distinguish them from original texts. We adopt this strategy as inspiration for learning a more robust representation from examples of negation but also paraphrasing. \citealt{kim2024finetuning} highlighted the limitation of CLIP at dealing with linguistic variations such as paraphrasing and their impact on tasks that require handling variations in user queries. They proposed ParaCLIP to address this problem by automatically generating two stages of paraphrasing from web-scale image captions using large language models (LLMs), first to convert captions into plain language and then to rephrase the result with similar vocabulary. By fine-tuning only the text encoder, they achieved substantial improvements on various retrieval tasks in comparison with CLIP. In this paper, we investigate negation together with paraphrasing. To the best of our knowledge, this is the first paper to study both challenges together. 

The above findings point to the benefit of text-image data augmentation for contrastive learning, while also highlighting the need to account for different lexical forms and semantic changes. We seek to achieve this by adding a new embedding projection to the CLIP architecture and a new loss function with pairwise components for negation and paraphrasing, respectively, with respect to the original text. The goal is to maintain performance as much as possible on the original image-text data while improving robustness to semantic variations. We refer to this architecture and associated training loss as SemCLIP. It extends CLIP by enabling joint learning of negation and paraphrasing. The two new loss function components added to the original contrastive loss of CLIP are: (1) \textit{paraphrasing loss}, \textit{$L_{paraphrase}$}, encouraging the mapping of paraphrased captions near to the original caption and corresponding image in embedding space, and (2) \textit{negation loss}, \textit{$L_{negation}$}, encouraging the mapping of negated captions far away from the original caption and corresponding image in embedding space. By jointly optimizing \textit{$L_{paraphrase}$} and \textit{$L_{negation}$} alongside the standard CLIP contrastive loss, \textit{$L_{contrastive}$}, results show that SemCLIP preserves top-1 image retrieval performance on original captions while achieving overall improved robustness (defined precisely in what follows using the mean-accuracy delta between original and negated captions) on downstream image classification tasks.

The remainder of the paper is organized as follows: related work is discussed in the next section. The SemCLIP architecture, proposed loss function and evaluation methodology are then introduced in Section \ref{methodology}. Experimental findings are presented and investigated in Section \ref{eval_experiments}, and followed by the conclusion and future work.

\section{Related Work} 
\label{sec:related_work}

The effectiveness of VLMs has been limited by a lack of semantic understanding. \citeauthor{yuksekgonul2023bow} (\citeyear{yuksekgonul2023bow}) showed that many VLMs behave as a \textit{bag-of-words}, overly sensitive to specific keywords. They introduced the Attributes, Relations and Orders (ARO) benchmark to evaluate attribute binding, relational understanding and sensitivity to the order of the words. The evaluation revealed near-chance performance even for state-of-the-art VLMs. To address the problem, \textit{composition-aware hard negative mining} was proposed. However, \citeauthor{hsieh2023sugarcrepe} (\citeyear{hsieh2023sugarcrepe}) later identified that generated negative captions in ARO were often flawed, grammatically incorrect or lacking a plausible visual context. To counter this vulnerability, \citeauthor{hsieh2023sugarcrepe} (\citeyear{hsieh2023sugarcrepe}) introduced \textit{Sugarcrepe}, intended to be a more robust benchmark dataset, that leveraged Large Language Models (LLMs) to generate more fluent and plausible negative captions, also applying adversarial refinements to mitigate annotation bias. Their findings revealed that the performance of many VLMs had been overstated, underscoring the complexity of accurately evaluating negation and compositionality in VLMs. Building on this work, Sugarcrepe\texttt{++} (SCPP) \citep{Dumpala2024-hx} extended the benchmark by incorporating further analysis of model sensitivity to both lexical and semantic variations.\footnote{Sugarcrepe focused on distinguishing between correct captions and lexically similar but semantically different negatives. It did not account for semantically-equivalent but lexically different paraphrasing.} Although we make no claims about compositionality in this paper, the datasets that are relevant for the evaluation of compositionality are also relevant here. SCPP associated each image with a triplet of captions: two semantically-equivalent but lexically different positive captions, and one hard negative caption. Evaluations using SCPP revealed that a strong performance on traditional benchmarks did not translate necessarily to success on this more fine-grained benchmark. Many models struggled with semantic coherence when object attributes or spatial relations were changed. Improving semantic invariance, i.e. robustness to lexical variation without change in semantics, remains a challenge and is one of the goals of this paper. The task is harder than investigating negation alone, as discussed in detail in our experimental evaluation. 

To address data augmentation asymmetry in CLIP, LaCLIP \citep{fan2023improving} used a random mix of original captions and diverse variations of LLM-generated \textit{rewrites}. ParaCLIP \citep{kim2024finetuning} introduced a more targeted paraphrasing generation using an LLM to clean up noisy captions, generating better paraphrasing data to finetune the text encoder. They showed that it is possible to enhance the performance of CLIP models by including retrieval with paraphrasing. LLip \citep{lavoie2024modeling} challenged CLIP’s assumption of a single image embedding, proposing instead a set of visual mixture tokens combined via multi-head attention to account for diverse captions. With the understanding that each sentence in a long caption may describe a partial aspect of the image, DreamLIP \citep{zheng2024dreamlip} sampled sub-captions to construct multiple positive pairs against image patches. This fine-grained alignment enabled better performance over CLIP without requiring significantly more data. LLip and DreamLIP showed the viability of associating an image with diverse textual descriptions. Our work takes inspiration from that work, although by contrast with LLip and DreamLIP, we seek to preserve CLIP’s modularity.

Another line of research has been focused on enforcing semantic exclusivity for negation, that is, training CLIP models to push apart embeddings that are semantically opposite even when they share high lexical similarity. CoN-CLIP \citep{singh2024learn} introduced the CC-Neg dataset which paired images with their captions and negated captions generated by an LLM. A modified contrastive loss seeks to separate image embeddings from those of negated captions. This inspired the formulation of a negation loss in this paper. Similarly, TripletCLIP \citep{Patel2024-ur} generated fully synthetic negative vision-language pairs. An LLM creates hard negative captions and a text-to-image model synthesizes corresponding negative images. Training sought to align positive image-caption pairs and push away negative pairs.

The above related work showed improvements over CLIP using either paraphrased captions or negated captions in isolation, sometimes showing substantial performance gains depending on the choice of datasets. Our work builds on such work but also addresses a harder problem: improvement across the semantically diverse space of paraphrasing and negation together. Our goal is also to maintain CLIP's modularity, adopting a simple embedding projection, and to obtain as little as possible degradation of performance of the original CLIP without negation or paraphrasing. We evaluate our approach on downstream image classification tasks in comparison with the closest related work investigating CLIP with negation alone \citep{singh2024learn} and with paraphrasing alone \citep{kim2024finetuning}.

\section{SemCLIP: Semantic CLIP with Paraphrasing and Negation}
\label{methodology}

We extend CLIP by proposing a new contrastive loss function that allows incorporating the learning of the concepts of paraphrasing and negation together into CLIP. This new representation, contrasting negation with the original captions but incorporating other ways of stating that original caption (paraphrasing), is expected to produce a more robust semantic alignment between text and image. By \textit{robust} we mean a representation capable of distinguishing small changes in text that may invert its meaning, as well as considerable changes in text that nevertheless maintain the original meaning, as measured on downstream tasks by the difference (the mean-accuracy delta) between standard accuracy and so-called negated accuracy, as defined below.

\begin{figure*}[htbp!]
  \centering
  \includegraphics[width=\linewidth,height=.28\textheight,keepaspectratio]{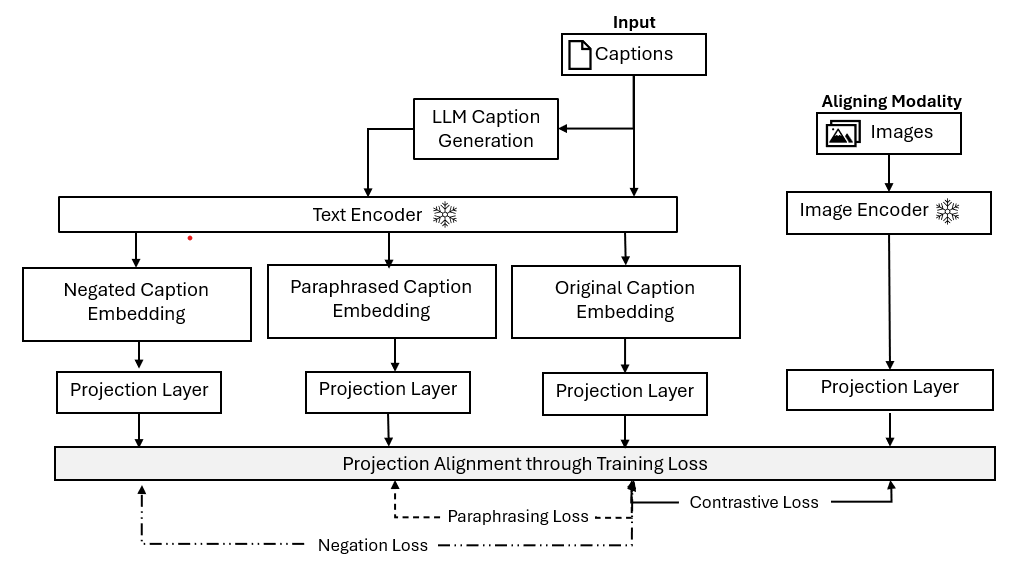}
  \caption{SemCLIP pipeline includes caption generation, fine-tuning of a pre-trained CLIP model, and downstream task. The SemCLIP architecture uses a contrastive loss $L_{contrastive}$, as usual, but also a paraphrasing loss $L_{paraphrase}$ and negation loss $L_{negation}$ to align the embeddings of image-text using a new pairwise embedding projection with training loss function given by Eq. \ref{eqn:total_train_loss} ({\tiny \SnowflakeChevron} denotes frozen models during training).}
  \label{fig:training_loss_scheme}
  \vspace{-12pt}
\end{figure*}

\noindent \textbf{Caption Augmentation:} given an image-text pair $(I, c)$, where $c$ is the textual caption of image $I$, SemCLIP generates two additional captions: (i) a paraphrased caption $c^+$, preserving the meaning of $c$ but differing in lexical or syntactic structure; (ii) a negated caption $c^-$, representing the opposite of the original description. Given a CLIP model, text embeddings can be obtained from $c$, $c^+$ and $c^-$, denoted respectively as $\mathbf{t}$, $\mathbf{t}^+$, and $\mathbf{t}^-$. All embeddings are $\ell_2$-normalized such that $\|\mathbf{t}\|_2 = \|\mathbf{t}^+\|_2 = \|\mathbf{t}^-\|_2 = 1$.

\noindent \textbf{Embedding Projections:} instead of disentangling trained embeddings to identify concepts and their negation, which has been shown to be very challenging \citep{higgins2018definitiondisentangledrepresentations}, we define a set of projection directions in embedding space and seek to learn to order embeddings based on those projections, aiming to separate $\mathbf{t^{+}}$ from $\mathbf{t^{-}}$. We initialize \( n \) orthonormal vectors
$
\mathbf{v}_i \in \mathbb{R}^d, \quad 1 \leq i \leq n, \quad
\|\mathbf{v}_i\|_2 = 1,
$
where \( d \) is the dimensionality of the embedding space. These vectors are sampled from a standard normal distribution and made orthonormal via QR decomposition. Given a unit-length embedding vector
$
\mathbf{t} \in \mathbb{R}^d,
$
we compute its scalar projection onto each \(\mathbf{v}_i\) by the dot product
$ 
p_i(\mathbf{t}) = \mathbf{t}^\top \mathbf{v}_i
$. The full projection of \(\mathbf{t}\) onto the subspace spanned by \(\{\mathbf{v}_i\}_{i=1}^n\) is represented by the vector of scalar projections: 
$
\mathbf{p}(\mathbf{t}) = \begin{bmatrix} p_1(\mathbf{t}), \dots,  p_n(\mathbf{t}) \end{bmatrix} \in \mathbb{R}^n
$.
We can alternatively express the projection by stacking the orthonormal vectors into a matrix
$
V = \begin{bmatrix} \mathbf{v}_1 & \mathbf{v}_2 & \cdots & \mathbf{v}_n \end{bmatrix} \in \mathbb{R}^{d \times n}.
$
The vector of scalar projections is given by the matrix multiplication:
$
\mathbf{p}(\mathbf{t}) = V^\top \mathbf{t} \in \mathbb{R}^n
$, where each component corresponds to
$
p_i(\mathbf{t}) = \mathbf{v}_i^\top \mathbf{t}
$. 
Optionally, \(\mathbf{p}(\mathbf{t})\) can be \(\ell_2\)-normalized to maintain consistent scale between projections, and projection vectors \(\{\mathbf{v}_i\}\) may be updated during training by gradient descent.

We consider two initializations. \textit{Random}: columns drawn from a standard normal distribution and orthonormalized as described above; \textit{Semantic} (for $n=1$): the single direction is warm-started to the empirical polarity direction $\mathbf{v} = (\bar{\mathbf{t}} - \bar{\mathbf{t}}^-)/\lVert \bar{\mathbf{t}} - \bar{\mathbf{t}}^- \rVert_2$, where $\bar{\mathbf{t}}$ and $\bar{\mathbf{t}}^-$ are the mean $\ell_2$-normalized embeddings of original and negated captions over a fixed held-out set of 256 caption pairs excluded from the training, validation and test sets.

The projection subspace dimension \(n\) is chosen to be much smaller than the original embedding dimension to encourage a more interpretable, low-dimensional and hopefully semantically meaningful representation. 
Rather than performing dimensionality reduction, our goal is to define a subspace where semantic relations for paraphrasing and negation correspond directly to geometric constraints on the projected embedding. Therefore, the projections onto the subspace are intended to provide a structured interpretation of the relationships between the two opposing types of captions, as follows. 

\noindent \textbf{Paraphrasing and Negation Losses:}
\label{loss_function}
Our training objective extends CLIP's original contrastive loss by introducing two new components: a \textit{paraphrasing loss} to enforce consistency across paraphrased captions, and a \textit{negation loss} to repel negated captions in projection space.
\noindent Let \(\{\mathbf{i}_i\}_{i=1}^N\) and \(\{\mathbf{t}_i\}_{i=1}^N\) denote batches of \(\ell_2\)-normalized image and text embeddings, respectively. The similarity scores between all image-text pairs in the batch are computed as: $S_{ij} = \tau \cdot \cos(\mathbf{i}_i, \mathbf{t}_j),$
where \(\tau = \exp(\theta)\) is a learnable temperature parameter, initialized from the pretrained checkpoint and, following standard CLIP practice, clamped to a maximum of 100 after each optimizer step.
The standard CLIP contrastive loss is then given by:
$
\mathcal{L}_{\text{contrastive}} = \frac{1}{2N} \sum_{i=1}^N \left( \mathcal{L}_{\mathrm{CE}}(S_{i,:}, i) + \mathcal{L}_{\mathrm{CE}}(S_{:,i}, i) \right),
$
where \(\mathcal{L}_{\mathrm{CE}}(\mathbf{v}, i)\) denotes the cross-entropy loss with logits \(\mathbf{v}\) and ground-truth class \(i\), while
$S_{i,:}$ and $S_{:,i}$
represent the \(i\)-th row and column of the similarity matrix \(S\), corresponding to image-to-text and text-to-image similarities, respectively. We define two additional loss-function components: (i) the \textit{paraphrasing loss} encourages the projections of an original caption and its paraphrase to lie in the same region of the subspace; (ii) the \textit{negation loss} encourages the projections of a caption and its negation to point in orthogonal or opposite directions:

\begin{equation}
     \mathcal{L}_{\text{paraphrase}} = 1 - \cos(\mathbf{p}(\mathbf{t}), \mathbf{p}(\mathbf{t}^+)), \quad
    \mathcal{L}_{\text{negation}} = \max\left(0, \cos(\mathbf{p}(\mathbf{t}), \mathbf{p}(\mathbf{t}^-)) \right),
\end{equation}
\noindent where $
\cos(\mathbf{a}, \mathbf{b}) = {\mathbf{a}^\top \mathbf{b}}/({\|\mathbf{a}\|_2 \|\mathbf{b}\|_2}). $

 For $n=1$, the cosine measure is identically $\pm 1$, a step
  function providing no gradient signal. In this case, we instead minimize a
  smooth soft-margin surrogate on the product of the scalar projections:
  \begin{equation}
      \mathcal{L}^{n=1}_{\text{paraphrase}} = \log\!\left(1 + e^{-\,p(\mathbf{t})\,p(\mathbf{t}^+)}\right), \quad
      \mathcal{L}^{n=1}_{\text{negation}} = \log\!\left(1 + e^{\,p(\mathbf{t})\,p(\mathbf{t}^-)}\right),
  \end{equation}
  which encourages paraphrase projections to simply share the sign of the original caption's projection and negation projections to take the opposite sign.
  
The losses encourage \(\mathbf{t}^+\) to project in the same direction as \(\mathbf{t}\) with cosine similarity close to 1, and \(\mathbf{t}^-\) to project in an orthogonal or opposing direction, with cosine similarity less than or equal to zero. 
The relation between original, paraphrased and negated captions is, therefore, captured via angular separation in this subspace. The combined loss function (Eq. \ref{eqn:total_train_loss}) aims to capture the essential semantic dichotomy between paraphrasing and negation through a simple geometric constraint. This is encoded at least in one dimension in the subspace when $n=1$, where paraphrases are encouraged to have the same sign as the original caption, and negations the opposite sign. The overall loss function is a weighted combination of all three components:
\begin{equation}
    \mathcal{L}_{\text{total}} = \frac{\alpha \mathcal{L}_{\text{contrastive}} + \beta \mathcal{L}_{\text{paraphrase}} + \gamma \mathcal{L}_{\text{negation}}}{\alpha + \beta + \gamma }
\label{eqn:total_train_loss}
\end{equation}

\noindent \textbf{Hyperparameters:} In the experiments to follow, for simplicity and ease of analysis, we use $\alpha, \beta, \gamma \in \{0,1\}$ only. Future work with a focus on the optimization of our results may consider a fine-grained grid search over $\alpha, \beta, \gamma$. Some additional hyperparameters control the behaviour of the projection mechanism: \textit{num\_projection\_vectors} (\textit{n}) is the number of projection directions and \textit{normalize\_projections} (Bool) is used to normalize w.r.t. $\ell_2$ the projections ($\mathbf{p(t)}$) prior to the calculation of the loss. For simplicity, all our experiments use \(n \in \{1, 2\}\). This should provide intuitive visualizations to help interpret or debug the learned projections, although we do not investigate this in detail in this paper. Exploring other values for \(n\) is of course possible and left as an interesting direction for future work to investigate how projection dimensionality affects semantic representation. We also evaluated projection normalization and projection polarity-basis initialisation in the experiments reported here (see Appendix \ref{apx:impl_details} for details). Finally, the projection vectors may be learnable, although our initial experiments did not show an improvement with learnable projections. See Appendix \ref{apx:ablation_studies} for ablation studies on these hyperparameters.
    
\noindent\textbf{Architecture:}
Figure \ref{fig:training_loss_scheme} illustrates the SemCLIP model and application of the combined loss function (Eq. \ref{eqn:total_train_loss}). The original CLIP architecture is composed of a vision encoder and a transformer-based text encoder. We augment it only by incorporating the shared projection space defined above. \textit{Vision Encoder:}
\label{vision_encoder}
we adopt a Vision Transformer (ViT-B/32) as the vision encoder, initialized with pre-trained weights from LAION-2B \citep{Cherti2023-dk}. Our code adapts the standard OpenCLIP implementation from \citealt{ilharco_gabriel_2021_5143773}. We use the pre-trained OpenCLIP models with frozen vision encoders during training. \textit{Text Encoder:}\label{text_encoder} we use a Transformer decoder with causal self-attention, learnable token embeddings and positional encoding. By default, it consists of $L=12$ layers with $H=8$ attention heads and uses a hidden dimension of $d=512$. The decoder operates over tokenized input captions, as usual, and the final output is derived from the representation at the end-of-sequence (EOS) token position. This representation is then passed through a linear projection layer to align with the dimensionality of the vision encoder output.

\section{Experimental Results}
\label{eval_experiments}

We evaluate SemCLIP trained on CC-Neg \citep{singh2024learn} and SCPP \citep{Dumpala2024-hx} because of the immediate related work, and we choose five datasets for the downstream zero-shot image classification task: CIFAR10 and CIFAR100 \citep{Krizhevsky2009CIFAR}, Flowers 102 \citep{Nilsback2008Flowers102}, Food 101 \citep{Bossard2014Food101} and Oxford-IIIT Pet \citep{Parkhi2012Pets}. CC-Neg has been used to evaluate negation understanding in vision-language models. Based on the CC-3M corpus, it contains 228,246 image-caption pairs accompanied by negated captions using terms such as ``no'', ``not'' and ``without''. We use LLMs to re-generate related captions but altering spatial relations, actions or states rather than inserting negation markers. See Synthetic Caption Generation below and Appendix~\ref{apx:prompt_templates} for details. SCPP is a much smaller but curated dataset used to evaluate compositional understanding of text-to-image mappings, with a total of 4,757 image-caption pairs where textual object attributes are swapped around but semantic similarity is maintained. Together, CC-Neg and SCPP offer an interesting mix for the purpose of our analysis, although to our knowledge no benchmark exists yet that is entirely suitable for the evaluation of the combination of negation and paraphrasing.

\begin{table*}[htbp!]
  \centering
  \caption{Evaluation of CLIP (baseline), CLIP with paraphrasing, CLIP with negation and SemCLIP (ours) on the CC-Neg and SCPP datasets. The table shows the average Top-1 accuracy on image-caption matching for the original captions for each model type. Top-1 accuracy remains comparable for SemCLIP.}  
 \label{tab:merged_performance_comparison}
  \begin{tabular}{l l c c c c}
    \toprule
\small Metric & \small Dataset &  \makecell{\small CLIP \\ \small Baseline} & \makecell{\small Paraphrase \\ \small only} & \makecell{\small Negation \\ \small only} & \makecell{\small SemCLIP \\ \small (ours)}\\
    \midrule
    \multirow{2}{*}{\small Original Caption (Top-1 Acc)} 
    & \small CC-Neg & \textbf{\small 33.7} & \small 31.9 & \small 31.2 & \small 31.1 \\
    & \small SCPP   & \small 66.4 & \small 67.6 & \small 68.7 & \textbf{\small 68.8} \\
    \bottomrule
  \end{tabular}
\end{table*}

\noindent\textbf{Synthetic Caption Generation:}
\label{syn_cap_generation}
To address the need for large quantities of paraphrased and negated captions, we employ a two-stage LLM-based synthetic caption generation pipeline. In the first stage, GPT 5.5 generates candidate paraphrased and negated captions from the original captions. In a second stage, Claude Opus 4.7 independently validates the captions by seeking to assess the quality of the generated captions. If a candidate caption fails validation, Claude Opus 4.7 provides a revised alternative. Using two independently developed models is intended to reduce correlated errors in generation and validation. The prompt templates for both stages are provided in Appendix \ref{apx:prompt_templates}. Synthetic caption generation is only applied to CCNeg, whereas the original captions with paraphrases and negations from SCPP are used directly. The pipeline is implemented using Azure Foundry Models for direct LLM invocation, with all model parameters kept at their default values during caption generation.

\noindent\textbf{Training Configuration:}
\label{train_config}
As detailed in the methodology section, image-caption pairs (together with their paraphrased and negated synthetic counterparts) are trained using contrastive learning. Training is carried out with the AdamW optimizer \citep{DBLP:journals/corr/abs-1711-05101}  ($\beta_{1}=0.9$, $\beta_{2}=0.98$, Weight decay: 0.2) for 200 epochs and at a learning rate of $5 \times 10^{-5}$ with a batch size of 1024. The learning rate follows a linear warm-up over 400 steps and then adopts a cosine annealing schedule \citep{loshchilov2017sgdrstochasticgradientdescent} without warm restarts. For SCPP, which is roughly 50 times smaller than CC-Neg, the batch size is reduced to 512 with a learning rate of $1 \times 10^{-4}$ and warm-up of 100 steps, so that each epoch retains a meaningful number of optimizer updates. For each run, the checkpoint with the highest validation score, computed as the mean of original-caption Top-1 accuracy, paraphrased-caption Top-1 accuracy and a scaled original-over-negation accuracy, is selected for test evaluation. The number of projection vectors, normalization of projections, the use of learnable projections and the polarity-basis initialisation options are also evaluated in an ablation test, with those results reported in Appendix \ref{apx:ablation_studies}. 

\label{expt_results}
Table \ref{tab:merged_performance_comparison} summarizes the results from 25 individual training runs\footnote{See Appendix B for details on the valid configurations used for the 25 training runs.} that systematically sweep the hyperparameters listed under Training Configuration. Four variants of CLIP were compared on the CC-Neg and SCPP datasets, each using different loss function terms, including the proposed SemCLIP model, as follows:
{\small
\begin{itemize}
  \item \textbf{CLIP Baseline} -- baseline model with only contrastive loss ($\beta=0$ and $\gamma=0$ in $\mathcal{L}_{\text{total}}$).
  \item \textbf{Paraphrase\_only} -- adds the paraphrase loss $L_{\text{paraphrase}}$ to the contrastive loss ($\gamma=0$ in $\mathcal{L}_{\text{total}}$).
  \item \textbf{Negation\_only} -- adds the negation loss $L_{\text{negation}}$ to the contrastive loss ($\beta=0$ in $\mathcal{L}_{\text{total}}$).
  \item \textbf{SemCLIP} -- adds both $L_{\text{paraphrase}}$ and $L_{\text{negation}}$ to the contrastive loss ($\alpha=\beta=\gamma=1$).
\end{itemize}
}

There are 22,799 held-out examples in the CC-Neg test set. Recovering the correct image (Top-1 accuracy) for approximately 1/3 of the cases, given the original caption, may appear to be low but it represents an exact match out of all other 22,798 images. As a sanity check, this indicates that SemCLIP's accuracy remains comparable to that of the other CLIP models despite the addition of the paraphrasing and negation terms to the loss function. Next, we turn to the evaluation of the main goal of SemCLIP: robustness on the downstream tasks.

\begin{figure*}[t]
    \centering
    \noindent
    \begin{minipage}{0.45\textwidth}
        \centering
        \includegraphics[width=\linewidth]{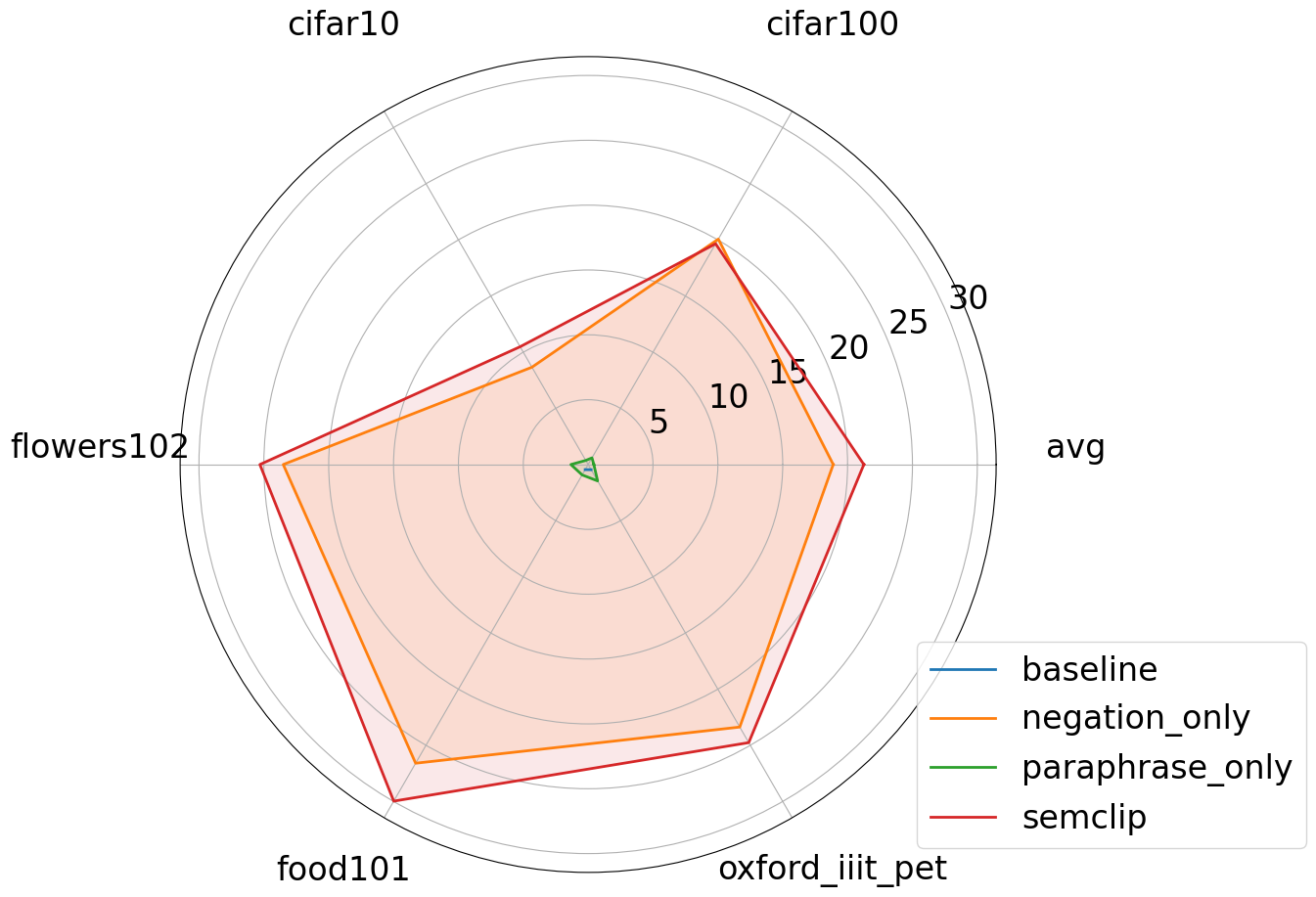}
        \par\vspace{5pt}
        \small (a) SemCLIP trained on CC-Neg.
        \label{fig:class_delta_ccneg_main}
    \end{minipage}
    \hfill
    \noindent
    \begin{minipage}{0.45\textwidth}
        \centering
        \includegraphics[width=\linewidth]{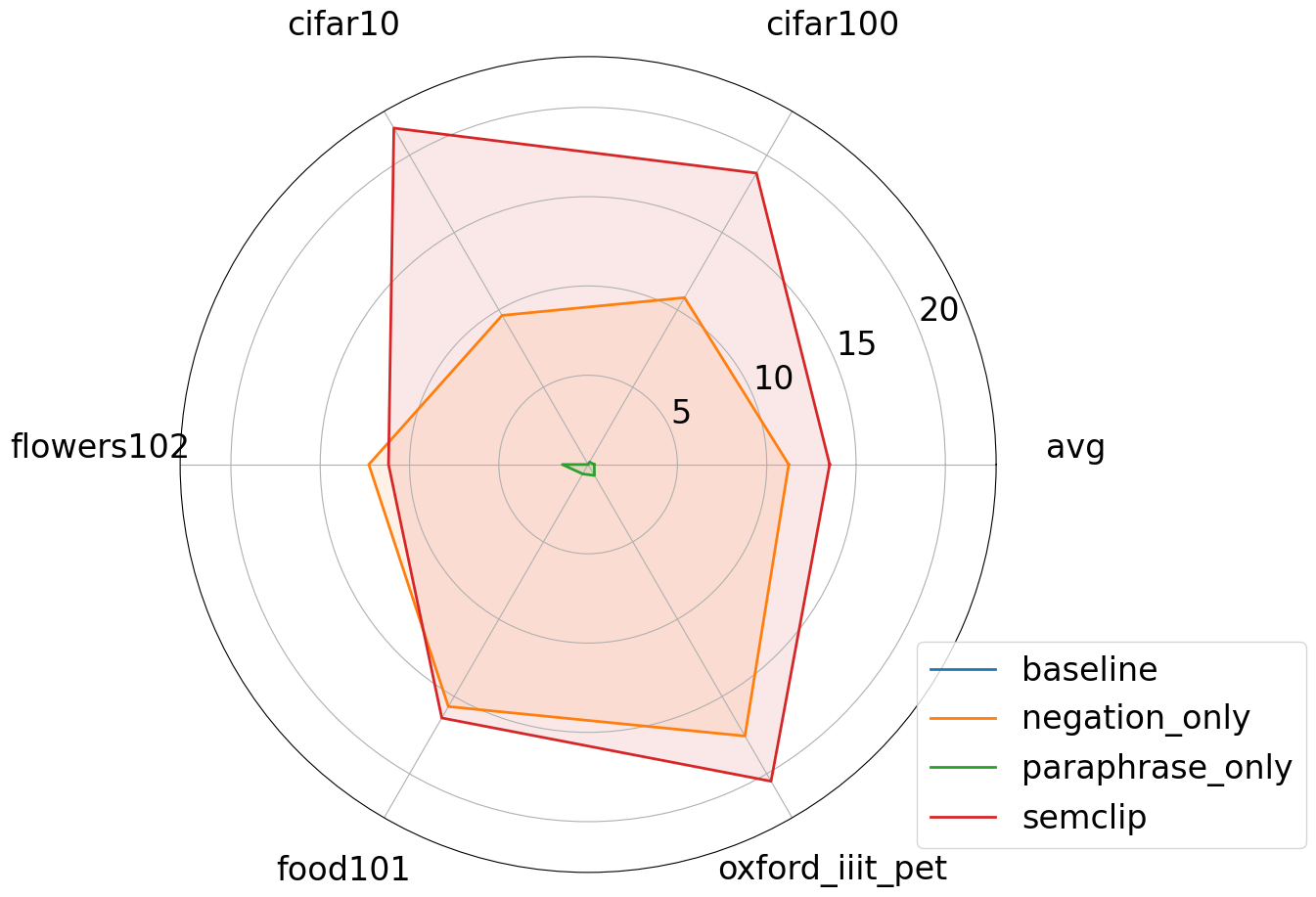}
        \par\vspace{5pt}
        \small (b) SemCLIP trained on SCPP.
        \label{fig:class_delta_scpp_main}
    \end{minipage}
    
    \caption{Downstream image classification task. Model robustness to negation is measured by the difference in accuracy (mean accuracy delta) between original and negated captions, e.g. ``This is not a photo of a \textless class\textgreater." In case a delta score is negative, it appears as a zero in the diagrams. SemCLIP achieves the highest delta on four out of five downstream classification datasets either when fine-tuned on CC-Neg (a) or SCPP (b); further detailed evaluations of the results are reported in Appendix \ref{apx:ablation_studies} (refer to Figures \ref{fig:downstream_class_acc_ccneg} and \ref{fig:downstream_class_acc_scpp} for a full comparison). }
    \label{fig:downstream_task_delta}
\end{figure*}

In \citealt{singh2024learn}, an original-over-negation accuracy score seeks to measure when an image is not matched to its negated caption. For each test image, the model is deemed correct when the cosine similarity to the original caption exceeds the similarity to its negated counterpart; original-over-negation accuracy is the fraction of test examples for which the original caption wins (chance level is 50\%).

We observe that direct comparisons are not possible since the task proposed here of handling both negation and paraphrasing is different from the tasks targeted by previous evaluations. Comparisons with the results reported in the literature should consider possible variations in implementation. Hence, all of the results reported in figures and tables here were obtained by running the OpenCLIP implementation from scratch, extended to include the paraphrasing loss, negation loss and SemCLIP. We nevertheless include for the sake of completeness below the results reported in the literature: based on the \textit{original-over-negation} results reported in \citep{singh2024learn} on the CC-Neg fine-tuning, SemCLIP achieves 82.0\% accuracy compared with CLIP's 65.70\%, Neg-CLIP's 62.63\%, FLAVA's 58.93\% \citep{Singh2021-yy} and BLIP's 62.31\% \citep{Li2022-ao}. SemCLIP only achieves a lower accuracy than CoN-CLIP's impressive 99.70\% \citep{singh2024learn}. To the best of our knowledge, CoN-CLIP has not been evaluated on SCPP.

We now evaluate the downstream zero-shot image classification tasks. For each of these downstream classification datasets, each class is represented by the text ``This is a photo of a \textless class\textgreater'' and by its negated counterpart ``This is not a photo of a \textless class\textgreater''. We evaluate on the training split of each downstream dataset; since none of the models is fine-tuned on these datasets, the evaluation is zero-shot for all of them and all models are compared on identical images. Given each image, each model predicts the class whose text embedding has the highest cosine similarity to the image embedding. Standard accuracy is the fraction of images whose predicted class matches the ground truth when using the original prompts; negated accuracy is the fraction matching the ground truth when using the negated prompts. We seek a high standard accuracy and a low negated accuracy. Hence, following \citealt{singh2024learn}, we report the difference between the two, known as the mean accuracy delta between standard and negated accuracies. A larger delta indicates an improved robustness to negation. As shown in Figure~\ref{fig:downstream_task_delta}, SemCLIP achieves the largest delta on four out of five downstream task datasets, regardless of CLIP fine-tuning with CC-Neg or SCPP. The individual standard and negated accuracy results are reported in Appendix Figures \ref{fig:downstream_class_acc_ccneg} and \ref{fig:downstream_class_acc_scpp}.

Appendix \ref{apx:ablation_studies} reports ablation results varying projection dimensions, making projections learnable and using different projection initialisations. The results indicate that the proposed joint learning of paraphrasing and negation makes it possible to obtain improved results, despite the challenges of the new learning task and with the use of a simple projection subspace. Performance improvements are not always guaranteed though. Exhaustive hyper-parameter tuning was not carried out and is left as future work, as our main goal was to study the combination of negation and paraphrasing in embedding space, how it might degrade performance (Table \ref{tab:merged_performance_comparison}) and improve robustness (Figure \ref{fig:downstream_task_delta}). Further experiments should follow, especially on real applications, ideally using well-curated data for the joint evaluation of negation and paraphrasing. 

\section{Conclusion and Future Work}
\label{conclusion}

We presented SemCLIP, which enhances CLIP's multimodal robustness to linguistic variations via joint learning of paraphrasing and negation, addressing well-known limitations of CLIP while seeking to preserve retrieval performance. More broadly, a neurosymbolic treatment of the task at hand should consider linguistic equivalence (building on paraphrasing), contradiction (building on various forms of negation), but also entailment in natural language inference \citep{srivastava2023imitationgamequantifyingextrapolating}. Incorporating all three within contrastive self-supervised objectives remains an open challenge for the field. While this paper does not tackle entailment directly, analysing negation and paraphrasing together should provide a more realistic setting for learning sound semantics in CLIP-like systems and for structuring embedding spaces around approximate equivalences, via paraphrasing, and contradictions, when opposing statements align in embedding spaces. This proposition is expected to yield neurosymbolic representations better suited for entailment inference, with future work aimed at extending the approach introduced here to more complex relationships such as non-verbal negation \citep{polarity}, and data augmentation strategies with multimodal LLMs as a way of realizing and interpreting entailment in contrastive models.

\clearpage
\bibliographystyle{iclr2026_conference}
\bibliography{references}

@article{higgins2018definitiondisentangledrepresentations,
      title={Towards a Definition of Disentangled Representations}, 
      author={Irina Higgins and David Amos and David Pfau and Sebastien Racaniere and Loic Matthey and Danilo Rezende and Alexander Lerchner},
      year={2018},
      eprint={1812.02230},
      archivePrefix={arXiv},
      primaryClass={cs.LG},
      url={https://arxiv.org/abs/1812.02230}, 
}

@article{fan2023improving,
	title = {Improving {CLIP} Training with Language Rewrites},
	url = {http://arxiv.org/abs/2305.20088},
	doi = {10.48550/arXiv.2305.20088},
	number = {{arXiv}:2305.20088},
	publisher = {{arXiv}},
	author = {Fan, Lijie and Krishnan, Dilip and Isola, Phillip and Katabi, Dina and Tian, Yonglong},
	urldate = {2025-02-04},
	date = {2023-10-28},
    year = {2023},
	langid = {english},
	eprinttype = {arxiv},
	eprint = {2305.20088 [cs]},
    archiveprefix = {arXiv},
}

@ARTICLE{Patel2024-ur,
  title         = "{TripletCLIP}: Improving compositional reasoning of {CLIP}
                   via synthetic vision-Language negatives",
  author        = "Patel, Maitreya and Kusumba, Abhiram and Cheng, Sheng and
                   Kim, Changhoon and Gokhale, Tejas and Baral, Chitta and Yang,
                   Yezhou",
  journal       = "arXiv [cs.CV] 2411.02545",
  month         =  nov,
  year          =  2024,
  url={https://arxiv.org/abs/2411.02545}, 
  archivePrefix = "arXiv",
  primaryClass  = "cs.CV"
}

@ARTICLE{polarity,  
AUTHOR={Vanek, Norbert  and Zhang, Haoruo },
TITLE={On truth and polarity in negation processing: language-specific effects in non-linguistic contexts},
JOURNAL={Frontiers in Psychology},
VOLUME={Volume 14 - 2023},
YEAR={2023},

DOI={10.3389/fpsyg.2023.1244249},
ISSN={1664-1078}}

@article{srivastava2023imitationgamequantifyingextrapolating,
      title={Beyond the Imitation Game: Quantifying and extrapolating the capabilities of language models}, 
      author={Aarohi Srivastava and Abhinav Rastogi and Abhishek Rao et al.},
      year={2023},
      eprint={2206.04615},
      archivePrefix={arXiv},
      primaryClass={cs.CL},
      url={https://arxiv.org/abs/2206.04615}, 
}

@article{3rdWave,
  title={Neurosymbolic {AI}: the 3rd wave},
  author={Artur S. d'Avila Garcez and L. Lamb},
  journal={Artificial Intelligence Review},
  year={2020},
  pages={1-20},
  url={https://api.semanticscholar.org/CorpusID:228083996}
}

@article{singh2024learn,
	title = {Learn ``No" to Say ``Yes" Better: Improving Vision-Language Models via Negations},
	url = {http://arxiv.org/abs/2403.20312},
	doi = {10.48550/arXiv.2403.20312},
	shorttitle = {Learn "No" to Say "Yes" Better},
	number = {{arXiv}:2403.20312},
	publisher = {{arXiv}},
	author = {Singh, Jaisidh and Shrivastava, Ishaan and Vatsa, Mayank and Singh, Richa and Bharati, Aparna},
	urldate = {2025-02-04},
	date = {2024-03-29},
    year = {2024},
	langid = {english},
	eprinttype = {arxiv},
	eprint = {2403.20312 [cs]},
    archiveprefix = {arXiv},
}

@article{zheng2024dreamlip,
	title = {{DreamLIP}: Language-Image Pre-training with Long Captions},
	url = {http://arxiv.org/abs/2403.17007},
	doi = {10.48550/arXiv.2403.17007},
	shorttitle = {{DreamLIP}},
	number = {{arXiv}:2403.17007},
	publisher = {{arXiv}},
	author = {Zheng, Kecheng and Zhang, Yifei and Wu, Wei and Lu, Fan and Ma, Shuailei and Jin, Xin and Chen, Wei and Shen, Yujun},
	urldate = {2025-02-04},
	date = {2024-03-25},
    year = {2024},
	langid = {english},
	eprinttype = {arxiv},
	eprint = {2403.17007 [cs]},
    archiveprefix = {arXiv},
}

@article{lavoie2024modeling,
	title = {Modeling Caption Diversity in Contrastive Vision-Language Pretraining},
	url = {http://arxiv.org/abs/2405.00740},
	doi = {10.48550/arXiv.2405.00740},
	number = {{arXiv}:2405.00740},
	publisher = {{arXiv}},
	author = {Lavoie, Samuel and Kirichenko, Polina and Ibrahim, Mark and Assran, Mahmoud and Wilson, Andrew Gordon and Courville, Aaron and Ballas, Nicolas},
	urldate = {2025-02-04},
	date = {2024-05-14},
    year = {2024},
	langid = {english},
	eprinttype = {arxiv},
	eprint = {2405.00740 [cs]},
    archiveprefix = {arXiv},
}

@article{yuksekgonul2023bow,
	title = {When and why vision-language models behave like bags-of-words, and what to do about it?},
	url = {http://arxiv.org/abs/2210.01936},
	doi = {10.48550/arXiv.2210.01936},
	publisher = {{arXiv}},
	author = {Yuksekgonul, Mert and Bianchi, Federico and Kalluri, Pratyusha and Jurafsky, Dan and Zou, James},
	urldate = {2025-02-14},
	date = {2023-03-23},
    year = {2023},
	langid = {english},
	eprinttype = {arxiv},
	eprint = {2210.01936 [cs]},
    archiveprefix = {arXiv},
}

@article{kim2024finetuning,
	title = {Fine-tuning {CLIP} Text Encoders with Two-step Paraphrasing},
	url = {http://arxiv.org/abs/2402.15120},
	doi = {10.48550/arXiv.2402.15120},
	number = {{arXiv}:2402.15120},
	publisher = {{arXiv}},
	author = {Kim, Hyunjae and Yoon, Seunghyun and Bui, Trung and Zhao, Handong and Tran, Quan and Dernoncourt, Franck and Kang, Jaewoo},
	urldate = {2025-02-14},
	date = {2024-02-23},
    year = {2024},
	langid = {english},
	eprinttype = {arxiv},
	eprint = {2402.15120 [cs]},
    archiveprefix = {arXiv},
}

@article{hsieh2023sugarcrepe,
	title = {{SugarCrepe}: Fixing Hackable Benchmarks for Vision-Language Compositionality},
	url = {http://arxiv.org/abs/2306.14610},
	doi = {10.48550/arXiv.2306.14610},
	shorttitle = {{SugarCrepe}},
	number = {{arXiv}:2306.14610},
	publisher = {{arXiv}},
	author = {Hsieh, Cheng-Yu and Zhang, Jieyu and Ma, Zixian and Kembhavi, Aniruddha and Krishna, Ranjay},
	urldate = {2025-04-14},
	date = {2023-06-26},
    year = {2023},
	langid = {english},
	eprinttype = {arxiv},
	eprint = {2306.14610 [cs]},
    archiveprefix = {arXiv},
}

@article{radford2021learning,
  title={Learning transferable visual models from natural language supervision},
  author={Radford, Alec and Kim, Jong Wook and Hallacy, Chris and Ramesh, Aditya and Goh, Gabriel and Agarwal, Sandhini and Sastry, Girish and Askell, Amanda and Mishkin, Pamela and Clark, Jack and others},
  journal={International Conference on Machine Learning},
  pages={8748--8763},
  year={2021},
  organization={PMLR}
}

@ARTICLE{Dumpala2024-hx,
      title={SUGARCREPE++ Dataset: Vision-Language Model Sensitivity to Semantic and Lexical Alterations}, 
      author={Sri Harsha Dumpala and Aman Jaiswal and Chandramouli Sastry and Evangelos Milios and Sageev Oore and Hassan Sajjad},
      year={2024},
      eprint={2406.11171},
      archivePrefix={arXiv},
      primaryClass={cs.CV},
      url={https://arxiv.org/abs/2406.11171}, 
}

@techreport{Krizhevsky2009CIFAR,
  title       = {Learning Multiple Layers of Features from Tiny Images},
  author      = {Alex Krizhevsky},
  institution = {University of Toronto},
  year        = {2009},
  note        = {Technical Report, CIFAR-10 and CIFAR-100 datasets},
  url         = {https://www.cs.toronto.edu/~kriz/learning-features-2009-TR.pdf}
}

@inproceedings{Bossard2014Food101,
  title     = {Food-101 -- Mining Discriminative Components with Random Forests},
  author    = {Lukas Bossard and Matthieu Guillaumin and Luc Van Gool},
  booktitle = {Proceedings of the European Conference on Computer Vision (ECCV)},
  year      = {2014},
  pages     = {446--461}
}

@inproceedings{Nilsback2008Flowers102,
  title     = {Automated Flower Classification over a Large Number of Classes},
  author    = {Maria-Elena Nilsback and Andrew Zisserman},
  booktitle = {Proceedings of the Indian Conference on Computer Vision, Graphics and Image Processing (ICVGIP)},
  year      = {2008},
  pages     = {722--729}
}

@inproceedings{Parkhi2012Pets,
  title     = {Cats and Dogs},
  author    = {Omkar M. Parkhi and Andrea Vedaldi and Andrew Zisserman and C. V. Jawahar},
  booktitle = {Proceedings of the IEEE Conference on Computer Vision and Pattern Recognition (CVPR)},
  year      = {2012},
  pages     = {3498--3505}
}

@INPROCEEDINGS{Cherti2023-dk,
  title     = "Reproducible scaling laws for contrastive language-image learning",
  author    = "Cherti, Mehdi and Beaumont, Romain and Wightman, Ross and
               Wortsman, Mitchell and Ilharco, Gabriel and Gordon, Cade and
               Schuhmann, Christoph and Schmidt, Ludwig and Jitsev, Jenia",
  booktitle = "2023 IEEE/CVF Conference on Computer Vision and Pattern
               Recognition (CVPR)",
  publisher = "IEEE",
  pages     = "2818--2829",
  month     =  jun,
  year      =  2023
}

@ARTICLE{ilharco_gabriel_2021_5143773,
  author       = {Ilharco, Gabriel and
                  Wortsman, Mitchell and
                  Wightman, Ross and
                  Gordon, Cade and
                  Carlini, Nicholas and
                  Taori, Rohan and
                  Dave, Achal and
                  Shankar, Vaishaal and
                  Namkoong, Hongseok and
                  Miller, John and
                  Hajishirzi, Hannaneh and
                  Farhadi, Ali and
                  Schmidt, Ludwig},
  title        = {Open{CLIP}},
  month        = jul,
  year         = 2021,
  publisher    = {Zenodo},
  doi          = {10.5281/zenodo.5143773},
  url          = {https://zenodo.org/records/7332113}
}

@article{DBLP:journals/corr/abs-1711-05101,
  author       = {Ilya Loshchilov and
                  Frank Hutter},
  title        = {Fixing Weight Decay Regularization in Adam},
  journal      = {CoRR},
  volume       = {abs/1711.05101},
  year         = {2017},
  url          = {http://arxiv.org/abs/1711.05101},
  eprinttype    = {arXiv},
  eprint       = {1711.05101},
  bibsource    = {dblp computer science bibliography, https://dblp.org}
}

@ARTICLE{Li2022-ao,
  title         = "{BLIP}: Bootstrapping language-image pre-training for unified
                   vision-language understanding and generation",
  author        = "Li, Junnan and Li, Dongxu and Xiong, Caiming and Hoi, Steven",
  url={https://arxiv.org/abs/2301.12597},
  journal       = "arXiv [cs.CV] 2301.12597",
  month         =  jan,
  year          =  2022,
  archivePrefix = "arXiv",
  primaryClass  = "cs.CV"
}

@ARTICLE{Singh2021-yy,
  title         = "{FLAVA}: A foundational language and vision alignment model",
  author        = "Singh, Amanpreet and Hu, Ronghang and Goswami, Vedanuj and
                   Couairon, Guillaume and Galuba, Wojciech and Rohrbach, Marcus
                   and Kiela, Douwe",
  journal       = "arXiv [cs.CV] 2112.04482",
  url={https://arxiv.org/abs/2112.04482},
  month         =  dec,
  year          =  2021,
  archivePrefix = "arXiv",
  primaryClass  = "cs.CV"
}

@article{loshchilov2017sgdrstochasticgradientdescent,
      title={SGDR: Stochastic Gradient Descent with Warm Restarts}, 
      author={Ilya Loshchilov and Frank Hutter},
      year={2017},
      eprint={1608.03983},
      archivePrefix={arXiv},
      primaryClass={cs.LG},
      url={https://arxiv.org/abs/1608.03983}, 
}

\clearpage
\appendix
\section{Prompt Templates for Caption Generation}
\label{apx:prompt_templates}
The following templates provide an example of generating paraphrased and negated captions using GPT-5.5 or Claude Opus 4.7 model as well as the subsequent quality validation by Claude Opus 4.7.

\begin{promptbox1}
\begin{lstlisting}[style=promptstyle]
Paraphrased caption candidate:
(GPT-5.5) - First Generation
Given this image caption: @@{caption}@@
Create a paraphrase that:
- Preserves the EXACT meaning and all details
- Uses different wording or sentence structure
- May reorder elements, use synonyms, or change voice (active/passive)
Provide only the paraphrased caption, nothing else.

(Claude Opus 4.7) - Regeneration or Fallback
Given this image caption: @@{caption}@@
Create a paraphrase that preserves the exact meaning but uses different wording.
Provide only the paraphrased caption, nothing else.

Negated caption candidate:
(GPT-5.5) - First Generation
Given this image caption: @@{caption}@@
Create a negation that contradicts the original by changing ONE of these:
1. Spatial relationships: under->next to, in front->behind, on->beside, in->out of
2. Actions: standing->sitting, walking->running, holding->dropping, playing->watching
3. States: driving->stuck, eating->spilling, looking->sleeping
Keep all other details identical. Make minimal changes.
Provide only the negated caption, nothing else.

(Claude Opus 4.7) - Regeneration or Fallback
Given this image caption: @@{caption}@@
Create a negation that contradicts the original by changing spatial relationships or actions.
Provide only the negated caption, nothing else.
\end{lstlisting}
\end{promptbox1}

\begin{promptbox2}
\begin{lstlisting}[style=promptstyle]
Paraphrased caption candidate
(Claude Opus 4.7)
The original sentence is: @@{original}@@
The generated paraphrase is: @@{generated}@@
Does the paraphrase accurately retain the meaning of the original? Respond with ONLY "Yes" or "No".

Negated caption candidate
(Claude Opus 4.7)
The original sentence is: @@{original}@@
The generated negation is: @@{generated}@@
Does the negation contradict or misrepresent the original? Respond with ONLY "Yes" or "No".
\end{lstlisting}
\end{promptbox2}

\clearpage
\section{Implementation Details}
\label{apx:impl_details}
Details of hardware used in this work are as follows:
\begin{enumerate}[label=\arabic*.]
    \item \textbf{Platform}: Azure (Standard NC40ads H100 v5)
    \item \textbf{vCPU}: 40
    \item \textbf{RAM}: 320GB
    \item \textbf{GPU}: 1 x Nvidia H100
    \item \textbf{Operating System}: Ubuntu 22.04
\end{enumerate}

Details of the training hyperparameters used are as follows:

\subsection{Image Normalization}
\textit{OpenAI} normalization values (as implemented via PyTorch transforms in the OpenCLIP library) were used for all training and testing runs.

\subsubsection*{Fixed Parameters}
\begin{enumerate}[label=\arabic*.]
    \item \textbf{Model}: ViT-B-32 (Pretrained: laion2b\_s34b\_b79k)
    \item \textbf{Epoch}: 200
    \item \textbf{Batch Size}: 1024 (CC-Neg) / 512 (SCPP)
    \item \textbf{Learning Rate}: 5e-5 (CC-Neg) / 1e-4 (SCPP)
    \item \textbf{Learning Rate schedule}: Linear warmup over 400 steps (CC-Neg) / 100 steps (SCPP); cosine annealing without warm restarts
    \item \textbf{Optimizer}: AdamW ($\beta_{1}$: 0.9, $\beta_{2}$: 0.98, Weight Decay: 0.2)
    \item \textbf{Precision}: bf16 mixed
    \item \textbf{Gradient clipping}: max norm 1.0 
    \item \textbf{Vision encoder}: frozen; the text encoder, the temperature parameter and, where applicable, the projection
  basis are trained.
    \item \textbf{Random Seed}: 42
\end{enumerate}

\subsubsection*{Varying Parameters}
\begin{enumerate}[resume, label=\arabic*.]
    \item \textbf{Number of Projection Vectors}: [1, 2]
    \item \textbf{Use Learnable Projections}: [True, False]
    \item \textbf{Projection Normalization}: [True, False]
    \item \textbf{Polarity Basis Initialisation}: [Random, Semantic]
\end{enumerate}

\subsubsection*{Valid Experimental Grid}
The experimental grid comprised of 6 binary parameters:
\begin{itemize} 
    \setlength\itemsep{0pt} 
    \item inclusion of the paraphrasing loss; 
    \item inclusion of the negation loss; 
    \item projection dimension \(n \in \{1,2\}\); 
    \item learnability of projection vectors; 
    \item use of projection normalization; and 
    \item polarity basis initialization (random/semantic). 
\end{itemize}

Although this gives (\(2^6\) = 64) nominal combinations, only valid and non-redundant configurations were evaluated. Projection normalization was excluded for (n=1), because normalizing a scalar projection collapses it to its sign; semantic initialization was used only for (n=1), since it defines a single polarity direction; and the CLIP baseline was run once because projection hyperparameters do not affect a model trained with only the standard contrastive loss. This leaves 25 distinct training configurations per fine-tuning dataset.

\clearpage
\section{Ablation Studies and Further Results}
\label{apx:ablation_studies}

Our first ablation study investigates the effect of key hyperparameters on image matching accuracies specifically: (1) the number of projection vectors, (2) the use of learnable projections, (3) the normalization of projection vectors, and (4) the
  polarity-basis initialisation. These factors are evaluated across three tasks: matching with original captions, paraphrased captions, and the original over negated caption task.

\begin{figure}[htbp!]
  \centering
  \noindent
  \begin{minipage}{\linewidth} 
    \centering
    \includegraphics[width=\linewidth,height=.23\textheight,keepaspectratio]{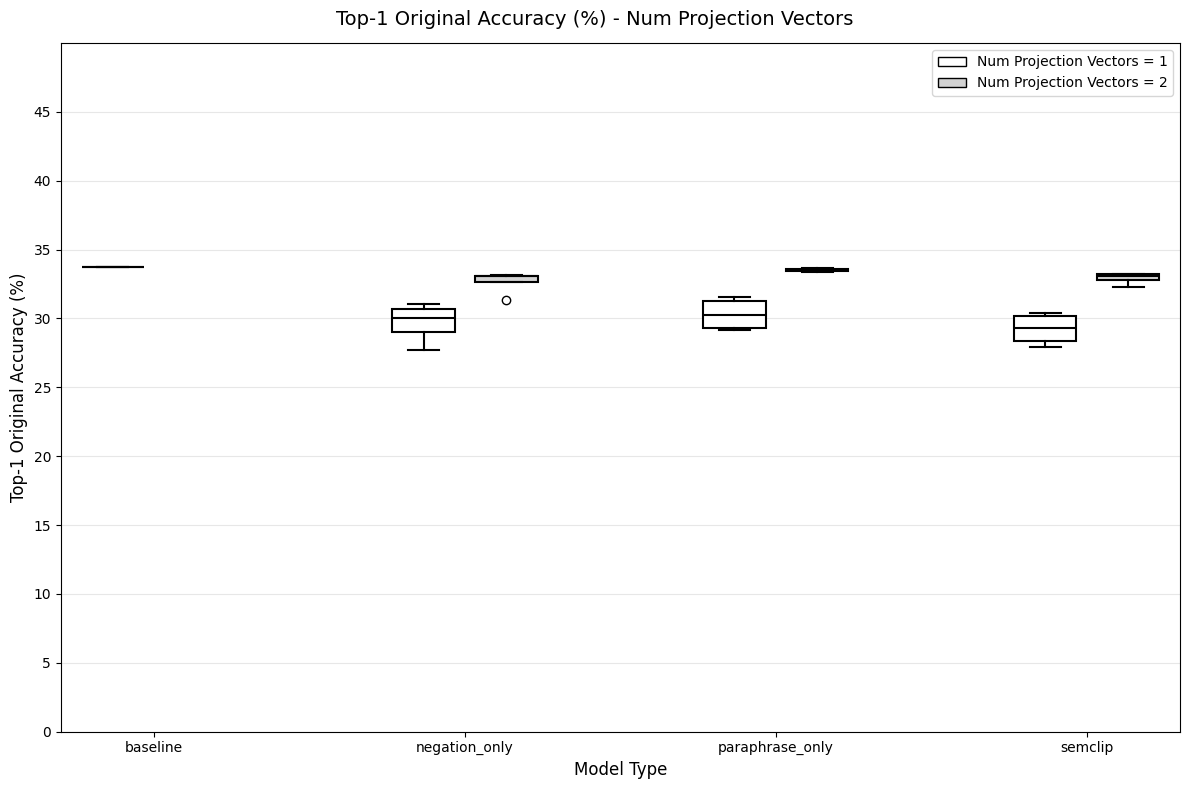}
    \par\vspace{5pt}
    \small (a) Effect of setting the number of projection vectors on Top-1 accuracy using original caption for image matching.
  \end{minipage}\par\vspace{10pt} 

  \noindent
  \begin{minipage}{\linewidth}
    \centering
    \includegraphics[width=\linewidth,height=.23\textheight,keepaspectratio]{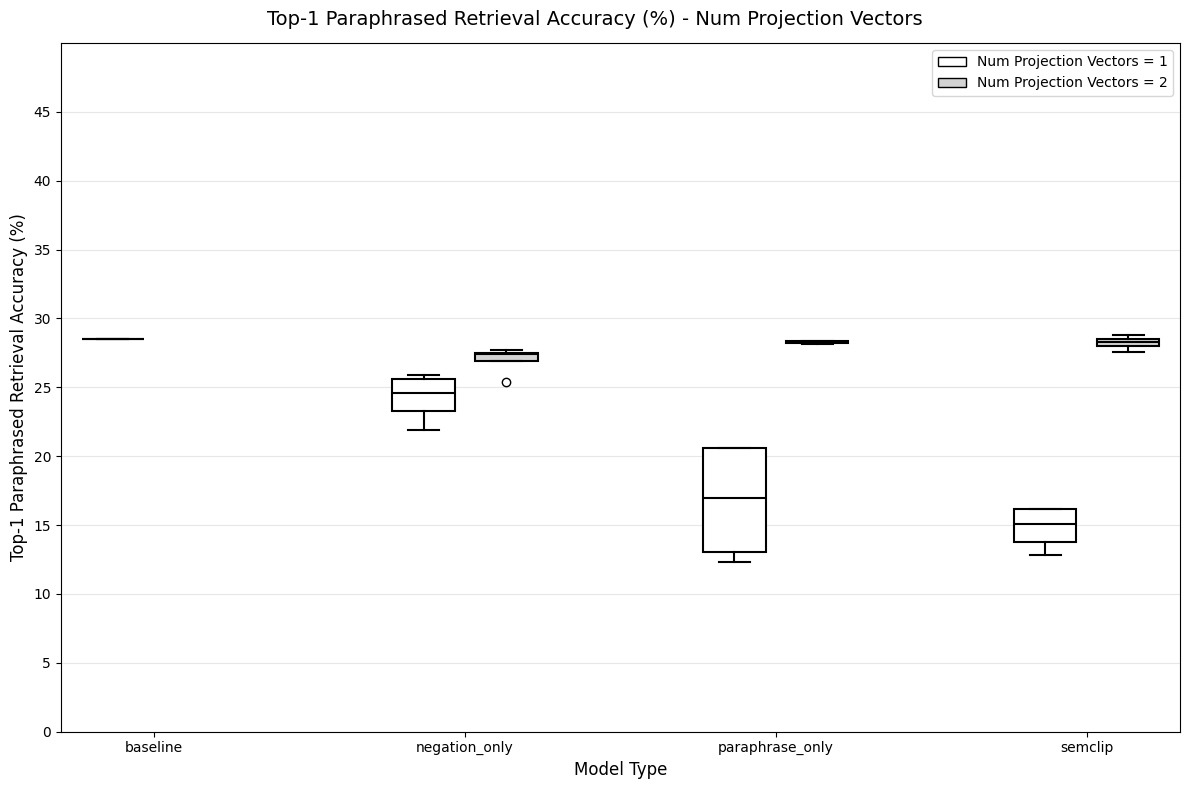}
    \par\vspace{5pt}
    \small (b) Effect of setting the number of projection vectors on Top-1 accuracy using paraphrased caption for image matching.
  \end{minipage}\par\vspace{10pt}

  \noindent
  \begin{minipage}{\linewidth}
    \centering
    \includegraphics[width=\linewidth,height=.23\textheight,keepaspectratio]{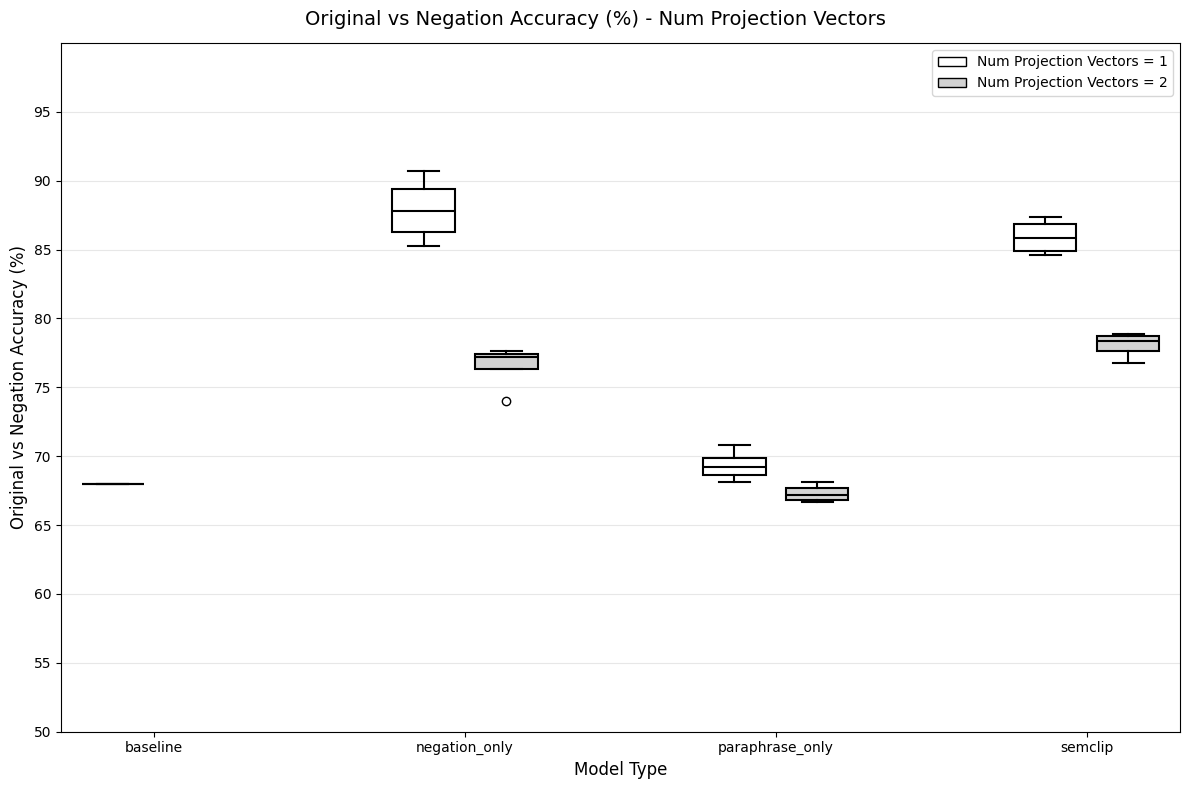}
    \par\vspace{5pt}
    \small (c) Effect of setting the number of projection vectors on Top-1 accuracy using original caption over negated caption for image matching.
  \end{minipage}

  \caption{Effect of setting the number of projection vectors on image matching accuracies using trained model (finetuned with CC-Neg dataset \citep{singh2024learn}).}
  \label{fig:hyperparameters_ablation_num_vec_ccneg}
\end{figure}

\begin{figure}[htbp!]
  \centering
  \noindent
  \begin{minipage}{\linewidth}
    \centering
    \includegraphics[width=\linewidth,height=.23\textheight,keepaspectratio]{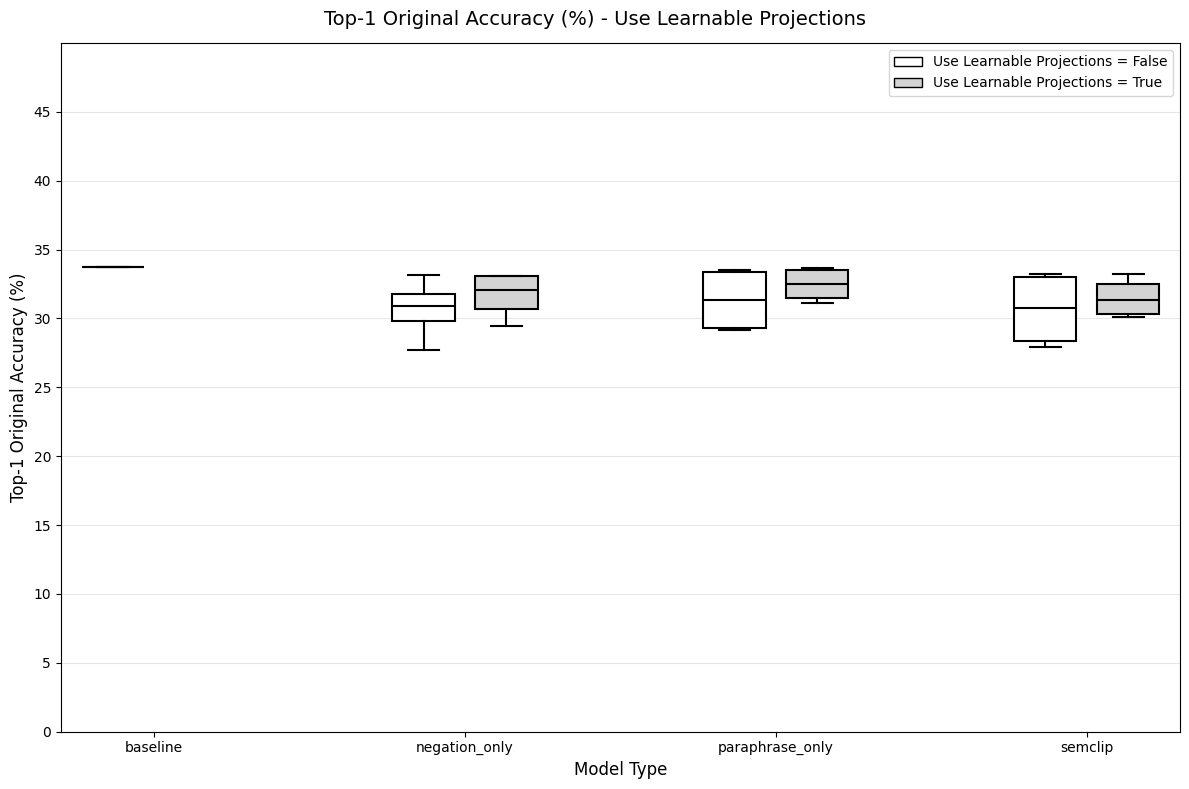}
    \par\vspace{5pt}
    \small (a) Effect of setting the use of learnable projection vectors on Top-1 accuracy using original caption for image matching.
  \end{minipage}\par\vspace{10pt}

  \noindent
  \begin{minipage}{\linewidth}
    \centering
    \includegraphics[width=\linewidth,height=.23\textheight,keepaspectratio]{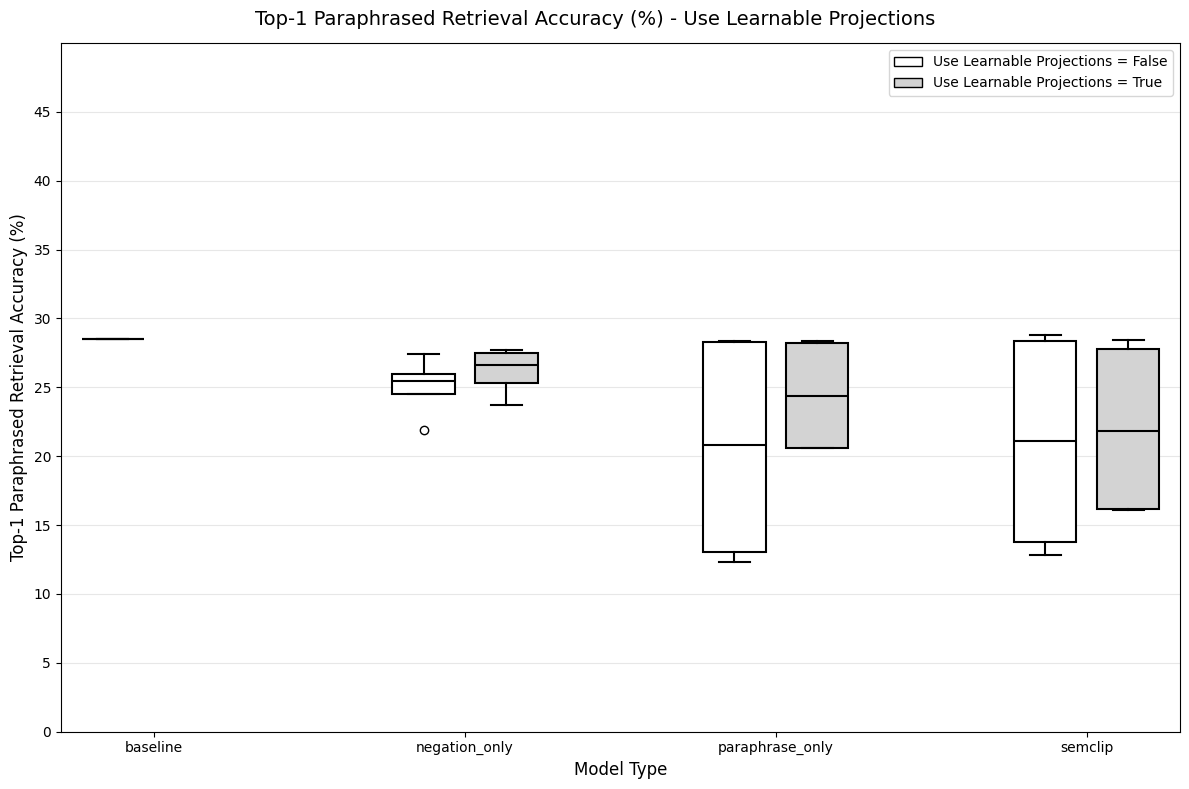}
    \par\vspace{5pt}
    \small (b) Effect of setting the use of learnable projection vectors on Top-1 accuracy using paraphrased caption for image matching.
  \end{minipage}\par\vspace{10pt}

  \noindent
  \begin{minipage}{\linewidth}
    \centering
    \includegraphics[width=\linewidth,height=.23\textheight,keepaspectratio]{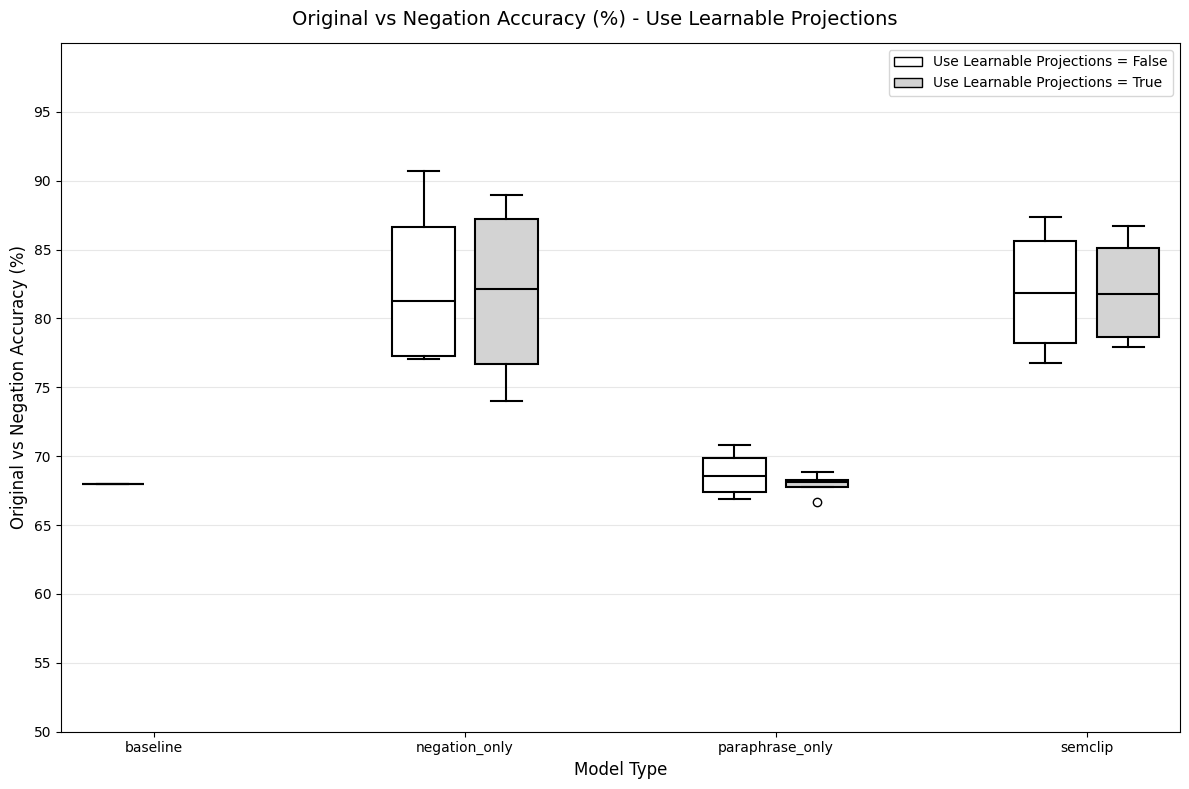}
    \par\vspace{5pt}
    \small (c) Effect of setting the use of learnable projection vectors on Top-1 accuracy using original caption over negated caption for image matching.
  \end{minipage}

  \caption{Effect of setting the use of learnable projection vectors on image matching accuracies using trained model (finetuned with CC-Neg dataset \citep{singh2024learn}).}
  \label{fig:hyperparameters_ablation_learn_proj_ccneg}
\end{figure}

\begin{figure}[htbp!]
  \centering
  \noindent
  \begin{minipage}{\linewidth}
    \centering
    \includegraphics[width=\linewidth,height=.23\textheight,keepaspectratio]{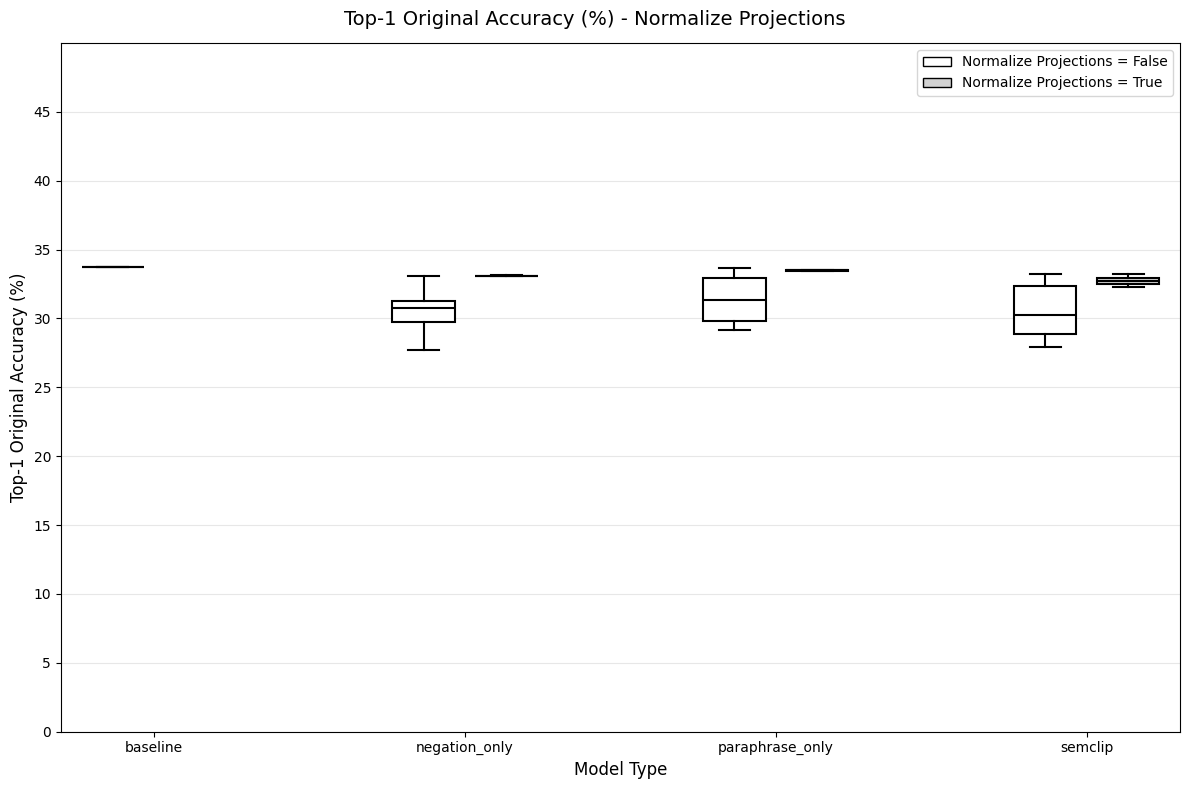}
    \par\vspace{5pt}
    \small (a) Effect of setting the projection vectors normalization on Top-1 accuracy using original caption for image matching.
  \end{minipage}\par\vspace{10pt}

  \noindent
  \begin{minipage}{\linewidth}
    \centering
    \includegraphics[width=\linewidth,height=.23\textheight,keepaspectratio]{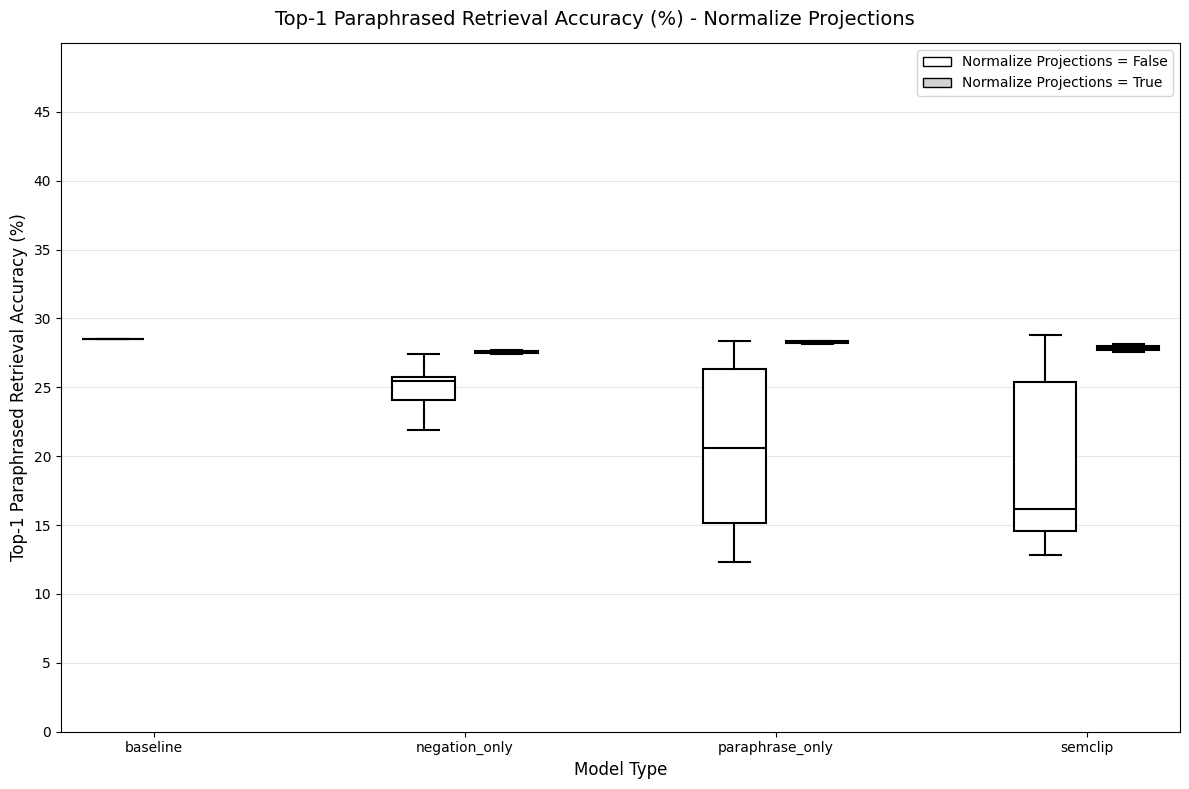}
    \par\vspace{5pt}
    \small (b) Effect of setting the projection vectors normalization on Top-1 accuracy using paraphrased caption for image matching.
  \end{minipage}\par\vspace{10pt}

  \noindent
  \begin{minipage}{\linewidth}
    \centering
    \includegraphics[width=\linewidth,height=.23\textheight,keepaspectratio]{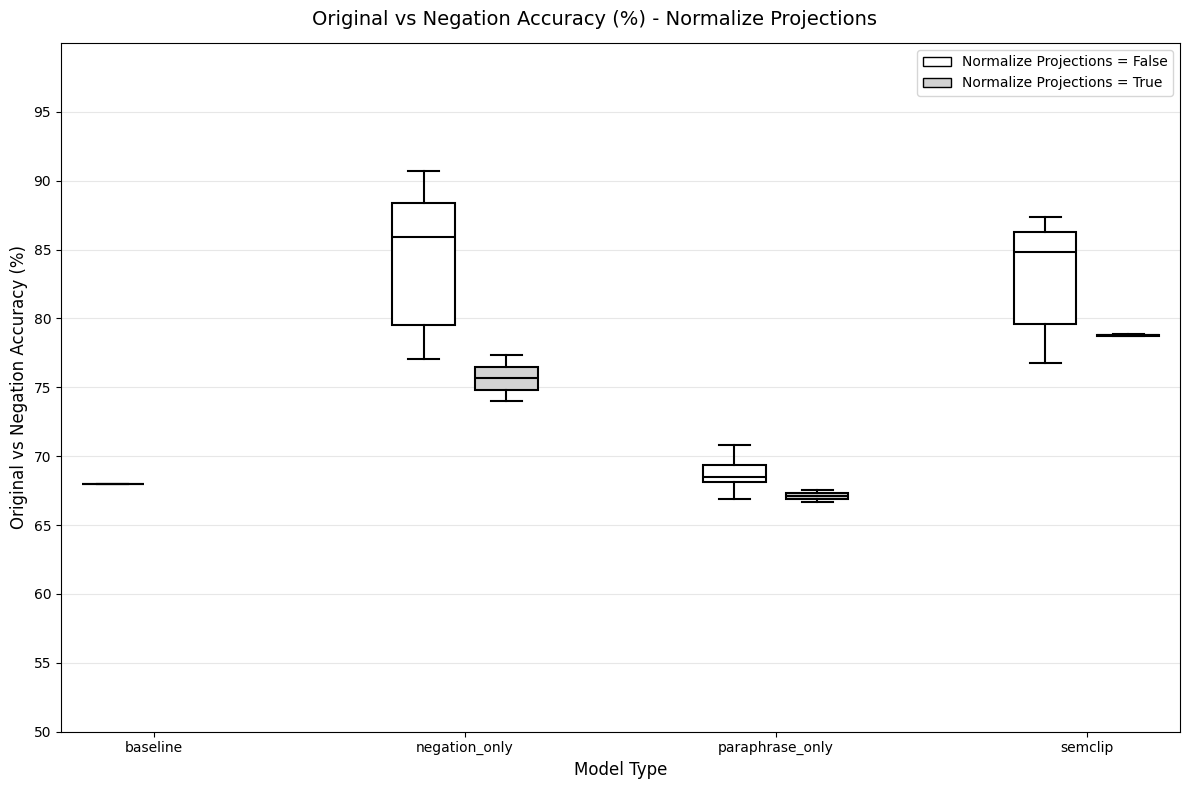}
    \par\vspace{5pt}
    \small (c) Effect of setting the projection vectors normalization on Top-1 accuracy using original caption over negated caption for image matching.
  \end{minipage}

  \caption{Effect of setting the projection vectors normalization on image matching accuracies using trained model (finetuned with CC-Neg dataset \citep{singh2024learn}).}
  \label{fig:hyperparameters_ablation_norm_proj_ccneg}
\end{figure}

\begin{figure}[htbp!]
  \centering
  \noindent
  \begin{minipage}{\linewidth}
    \centering
    \includegraphics[width=\linewidth,height=.23\textheight,keepaspectratio]{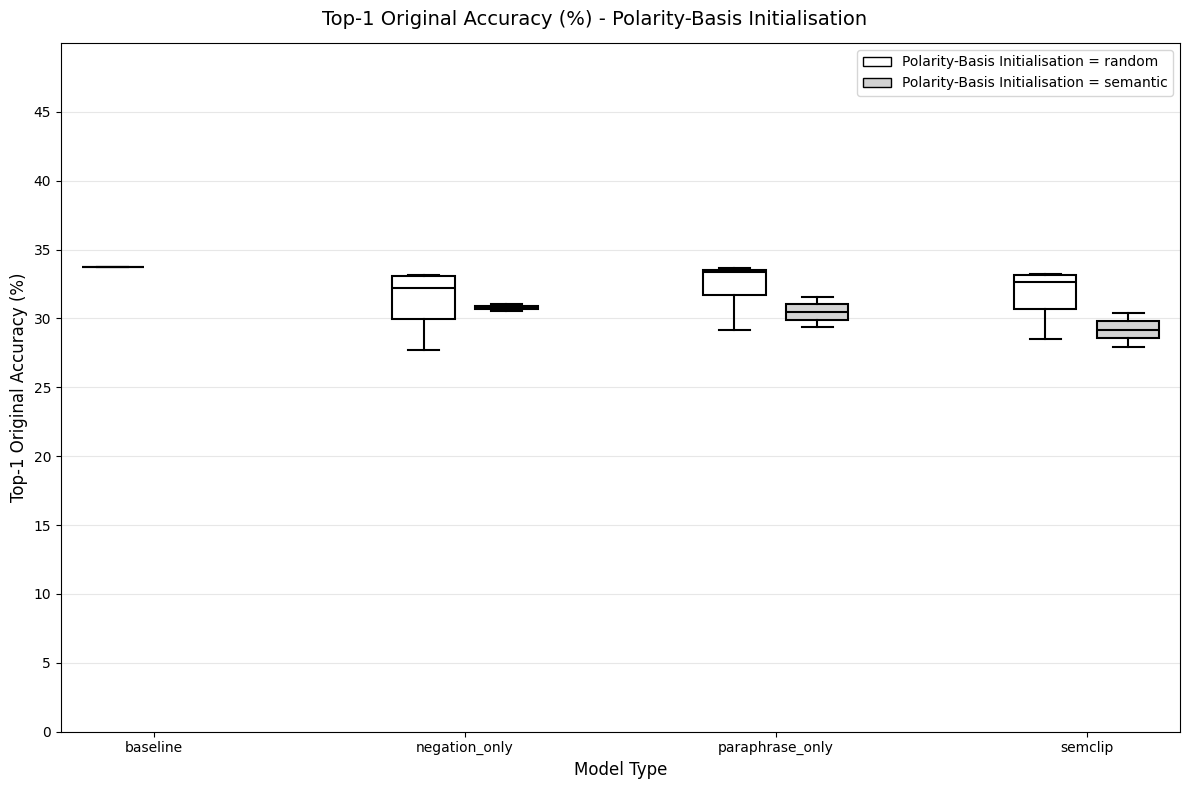}
    \par\vspace{5pt}
    \small (a) Effect of setting the polarity-basis initialisation on Top-1 accuracy using original caption for image matching.
    \label{fig:hp_ablation_basis_init_orig_ccneg}
  \end{minipage}\par\vspace{10pt}

  \noindent
  \begin{minipage}{\linewidth}
    \centering
    \includegraphics[width=\linewidth,height=.23\textheight,keepaspectratio]{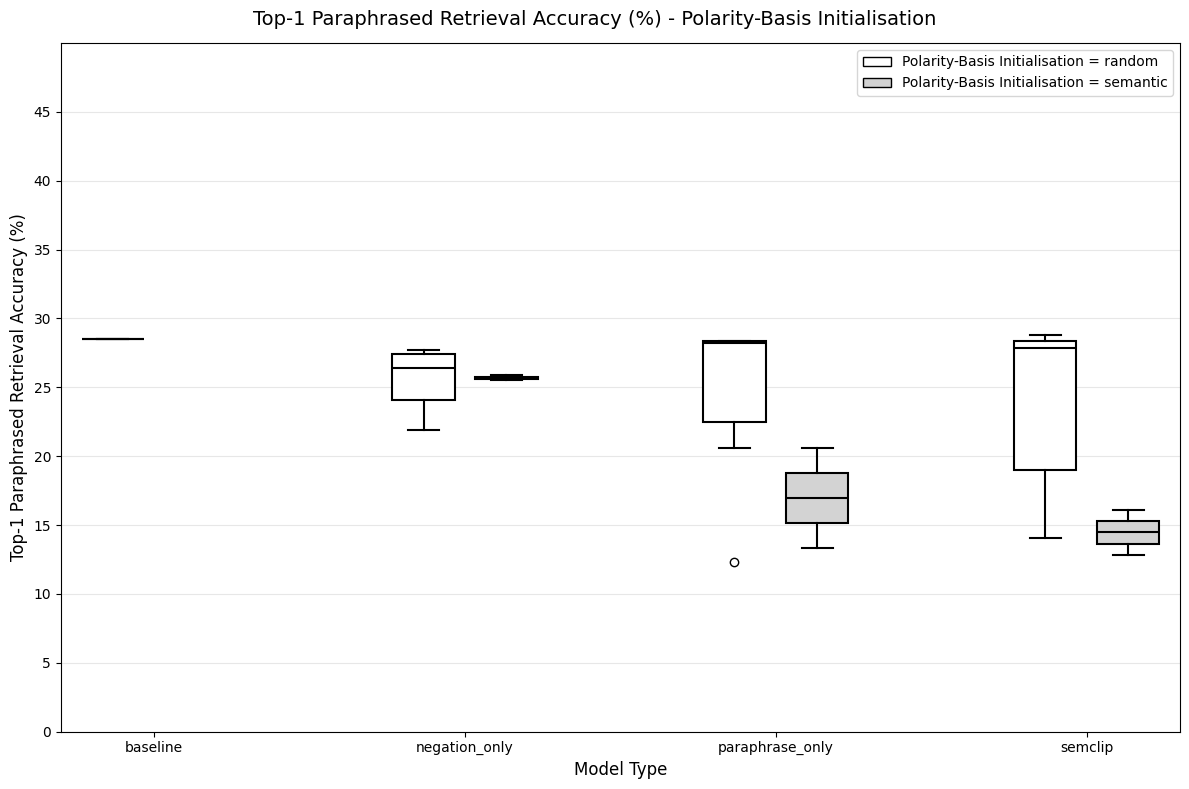}
    \par\vspace{5pt}
    \small (b) Effect of setting the polarity-basis initialisation on Top-1 accuracy using paraphrased caption for image matching.
    \label{fig:hp_ablation_basis_init_para_ccneg}
  \end{minipage}\par\vspace{10pt}

  \noindent
  \begin{minipage}{\linewidth}
    \centering
    \includegraphics[width=\linewidth,height=.23\textheight,keepaspectratio]{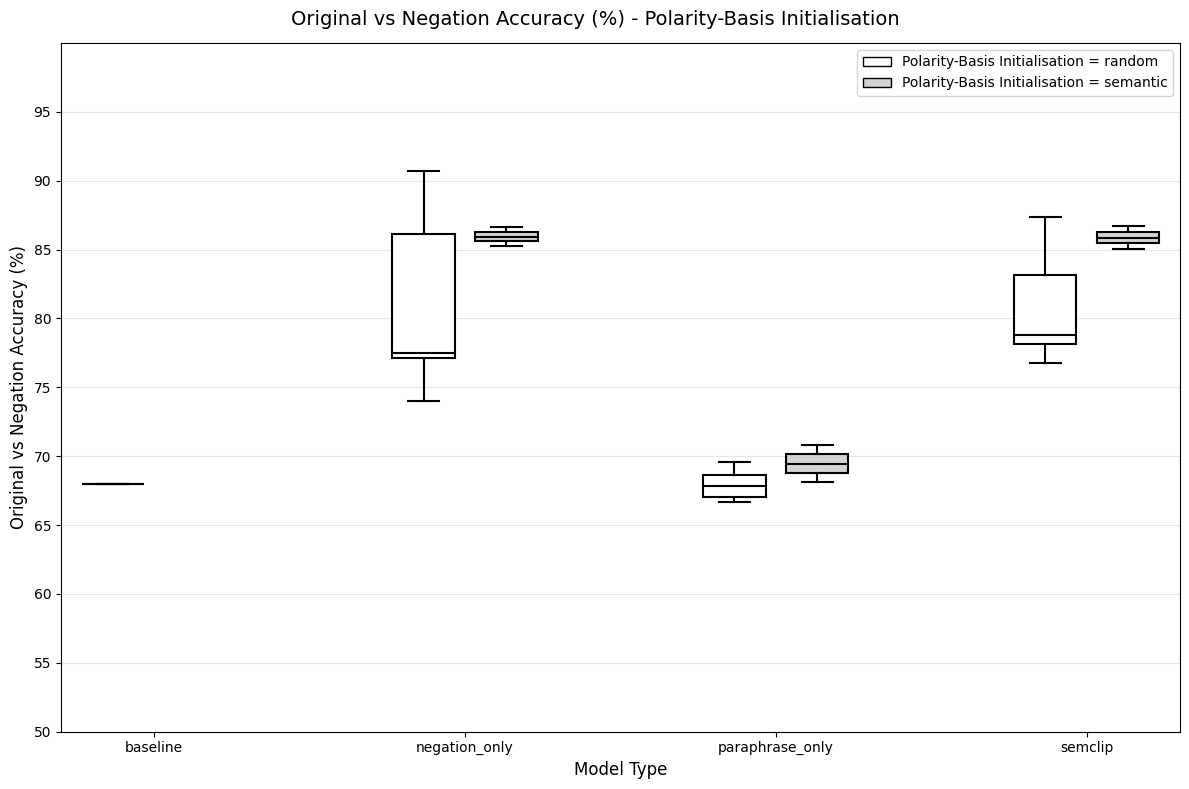}
    \par\vspace{5pt}
    \small (c) Effect of setting the polarity-basis initialisation on Top-1 accuracy using original caption over negated caption for image matching.
    \label{fig:hp_ablation_basis_init_oon_ccneg}
  \end{minipage}

  \caption{Effect of setting the polarity-basis initialisation on image matching accuracies using trained model (finetuned with CC-Neg dataset \citep{singh2024learn}).}
  \label{fig:hyperparameters_ablation_basis_init_ccneg}
\end{figure}

Figures~\ref{fig:hyperparameters_ablation_num_vec_ccneg}(a), \ref{fig:hyperparameters_ablation_learn_proj_ccneg}(a), \ref{fig:hyperparameters_ablation_norm_proj_ccneg}(a) and \ref{fig:hyperparameters_ablation_basis_init_ccneg}(a) present results using original captions. All models trained with CC-Neg dataset \citep{singh2024learn}, including our proposed SemCLIP, achieve comparable accuracies to the CLIP baseline, consistently at around 27.7\% - 33.7\%. The selected hyperparameters show negligible impact, indicating that our modifications preserve the model's core image matching capability using original captions.

By contrast, performance applied to paraphrased captions (Figures~\ref{fig:hyperparameters_ablation_num_vec_ccneg}(b), \ref{fig:hyperparameters_ablation_learn_proj_ccneg}(b), \ref{fig:hyperparameters_ablation_norm_proj_ccneg}(b) and \ref{fig:hyperparameters_ablation_basis_init_ccneg}(b)) drops, ranging between 12.3\% and 28.8\%. This decrease highlights the challenge posed by linguistic variation. Among the hyperparameters, the number of projection vectors has the most noticeable influence on SemCLIP with a 2D projection producing higher accuracy than a 1D projection.

The most pronounced hyperparameter effects appear in the original over negated task. As presented in Figures~\ref{fig:hyperparameters_ablation_num_vec_ccneg}(c), \ref{fig:hyperparameters_ablation_learn_proj_ccneg}(c), \ref{fig:hyperparameters_ablation_norm_proj_ccneg}(c) and \ref{fig:hyperparameters_ablation_basis_init_ccneg}(c), the CLIP baseline and Paraphrase\_only models struggle with accuracies below 70.8\%, suggesting an inability to reliably distinguish between the intended caption and its negation. By contrast, models explicitly trained on negation data (Negation\_only and SemCLIP) perform substantially better. SemCLIP achieves peak accuracy of 87.4\%. Other hyperparameters have limited impact on this task across all models.

\begin{figure}[htbp!]
  \centering
  \noindent
  \begin{minipage}{\linewidth}
    \centering
    \includegraphics[width=\linewidth,height=.23\textheight,keepaspectratio]{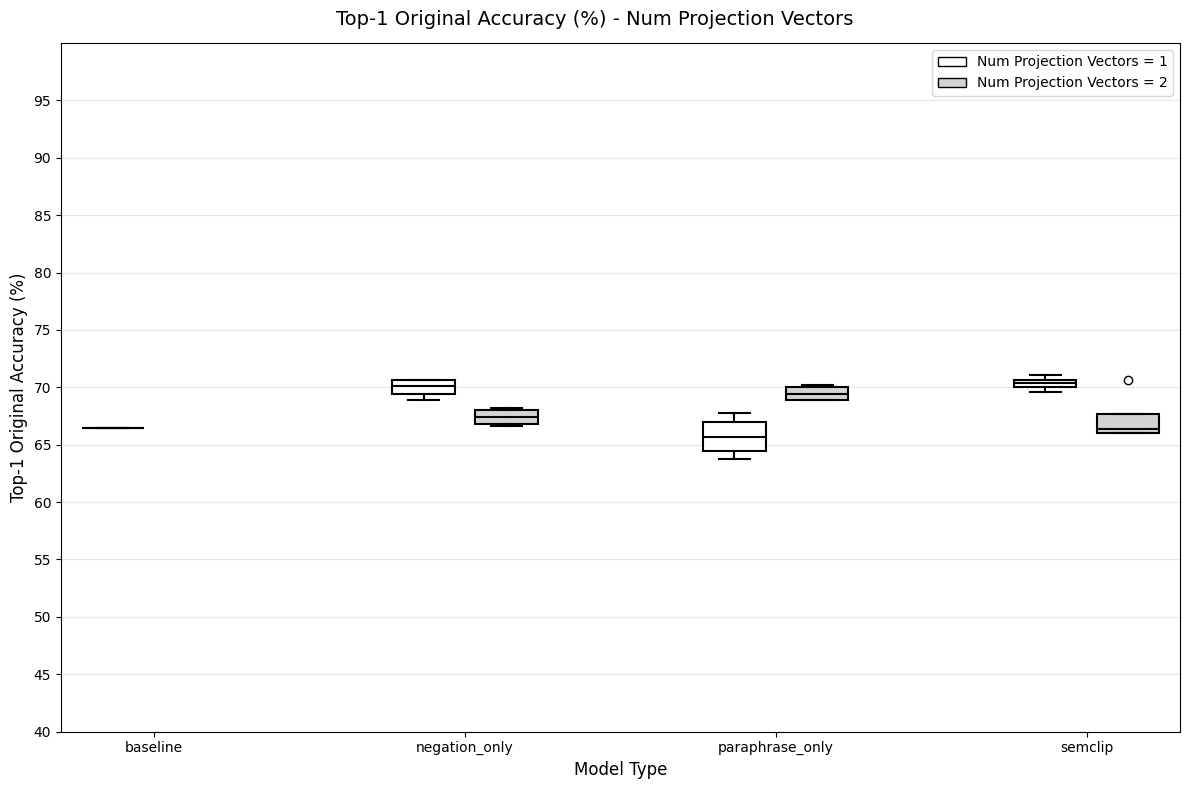}
    \par\vspace{5pt}
    \small (a) Effect of setting the number of projection vectors on Top-1 accuracy using original caption for image matching.
    \label{fig:hp_ablation_num_vec_orig_scpp}
  \end{minipage}\par\vspace{10pt}

  \noindent
  \begin{minipage}{\linewidth}
    \centering
    \includegraphics[width=\linewidth,height=.23\textheight,keepaspectratio]{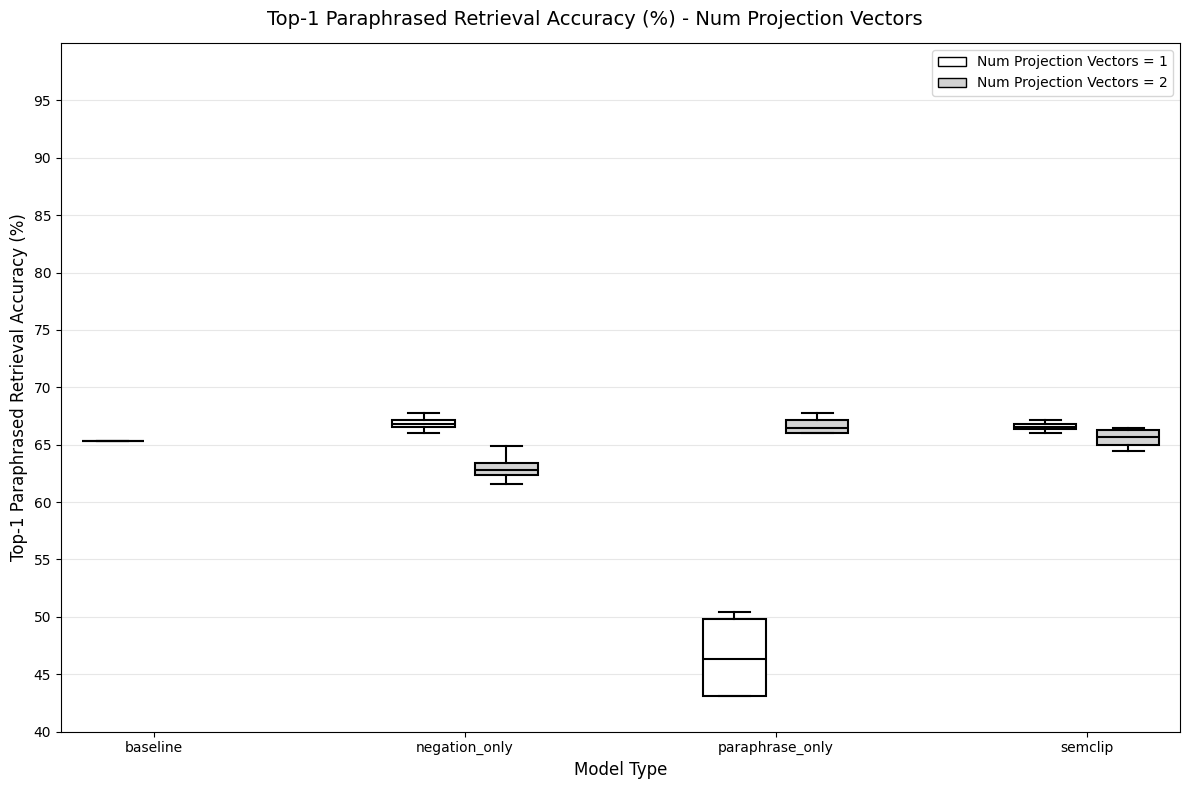}
    \par\vspace{5pt}
    \small (b) Effect of setting the number of projection vectors on Top-1 accuracy using paraphrased caption for image matching.
    \label{fig:hp_ablation_num_vec_para_scpp}
  \end{minipage}\par\vspace{10pt}

  \noindent
  \begin{minipage}{\linewidth}
    \centering
    \includegraphics[width=\linewidth,height=.23\textheight,keepaspectratio]{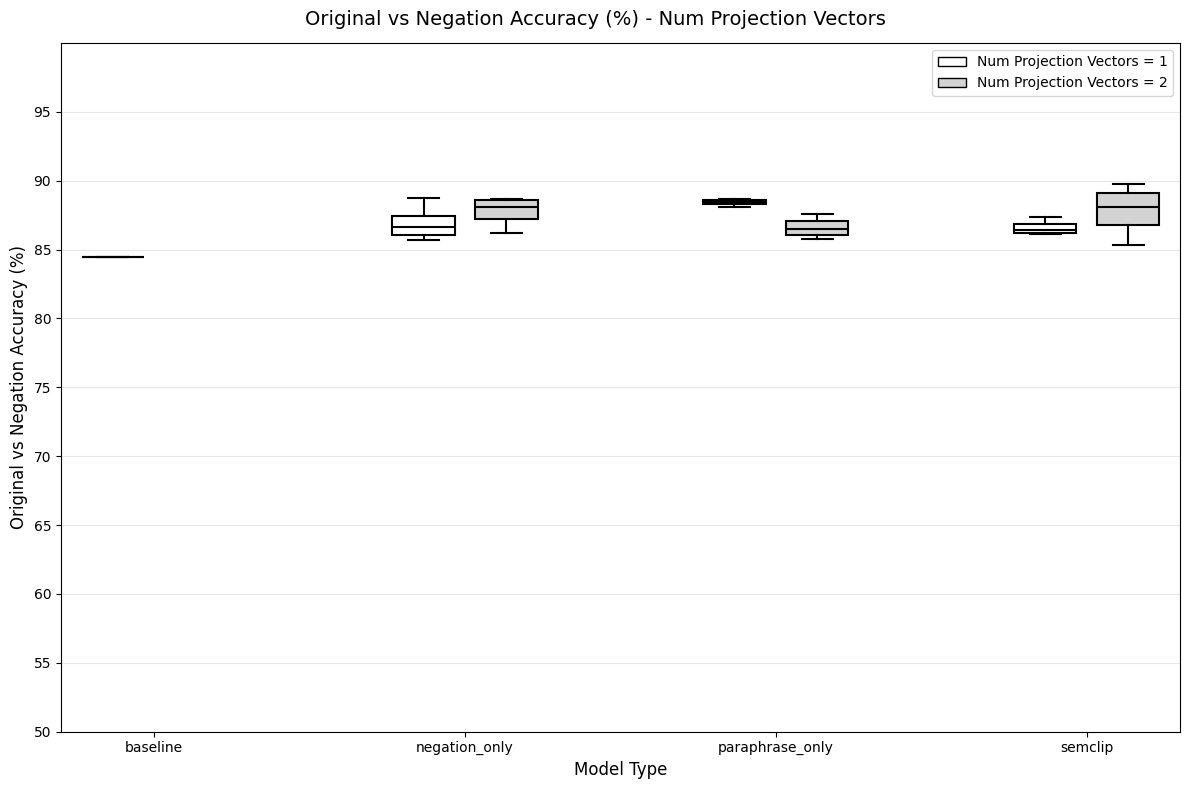}
    \par\vspace{5pt}
    \small (c) Effect of setting the number of projection vectors on Top-1 accuracy using original caption over negated caption for image matching.
    \label{fig:hp_ablation_num_vec_oon_scpp}
  \end{minipage}

  \caption{Effect of setting the number of projection vectors on image matching accuracies using trained model (finetuned with Sugarcrepe\texttt{++} dataset \citep{Dumpala2024-hx}).}
  \label{fig:hyperparameters_ablation_num_vec_scpp}
\end{figure}

\begin{figure}[htbp!]
  \centering
  \noindent
  \begin{minipage}{\linewidth}
    \centering
    \includegraphics[width=\linewidth,height=.23\textheight,keepaspectratio]{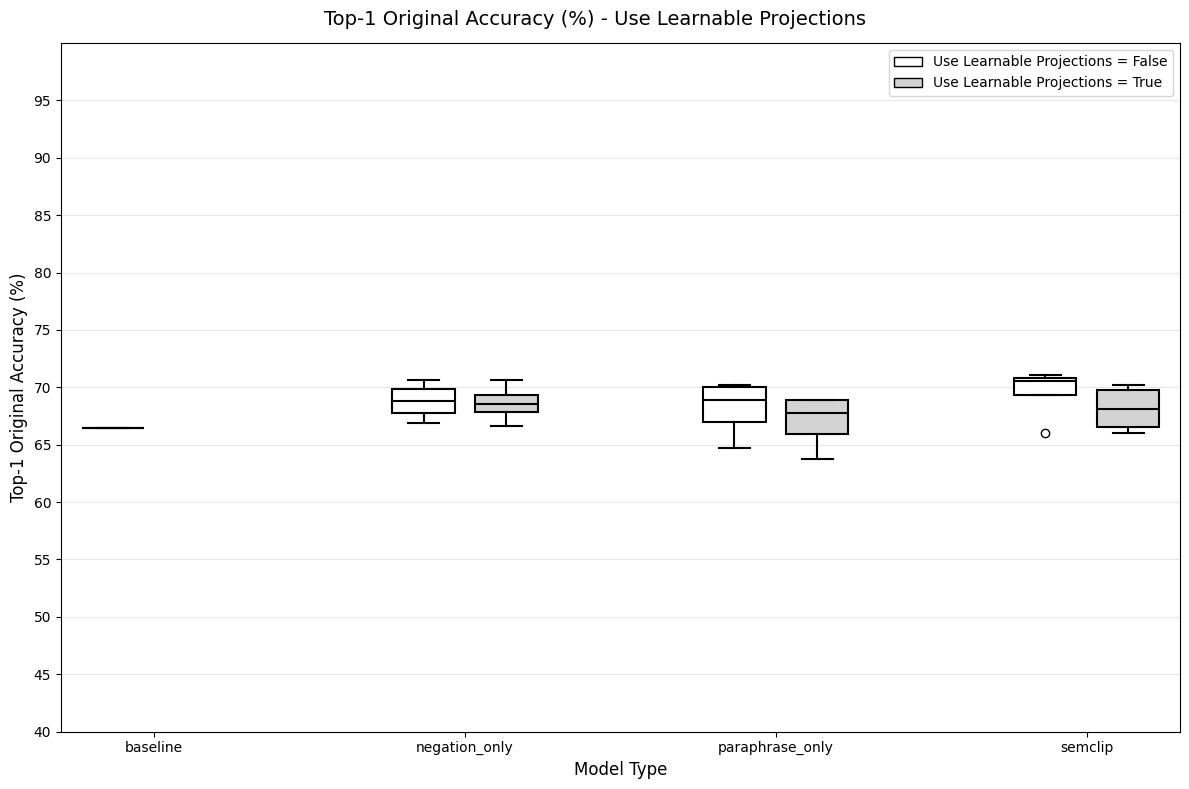}
    \par\vspace{5pt}
    \small (a) Effect of setting the use of learnable projection vectors on Top-1 accuracy using original caption for image matching.
    \label{fig:hp_ablation_learn_proj_orig_scpp}
  \end{minipage}\par\vspace{10pt}

  \noindent
  \begin{minipage}{\linewidth}
    \centering
    \includegraphics[width=\linewidth,height=.23\textheight,keepaspectratio]{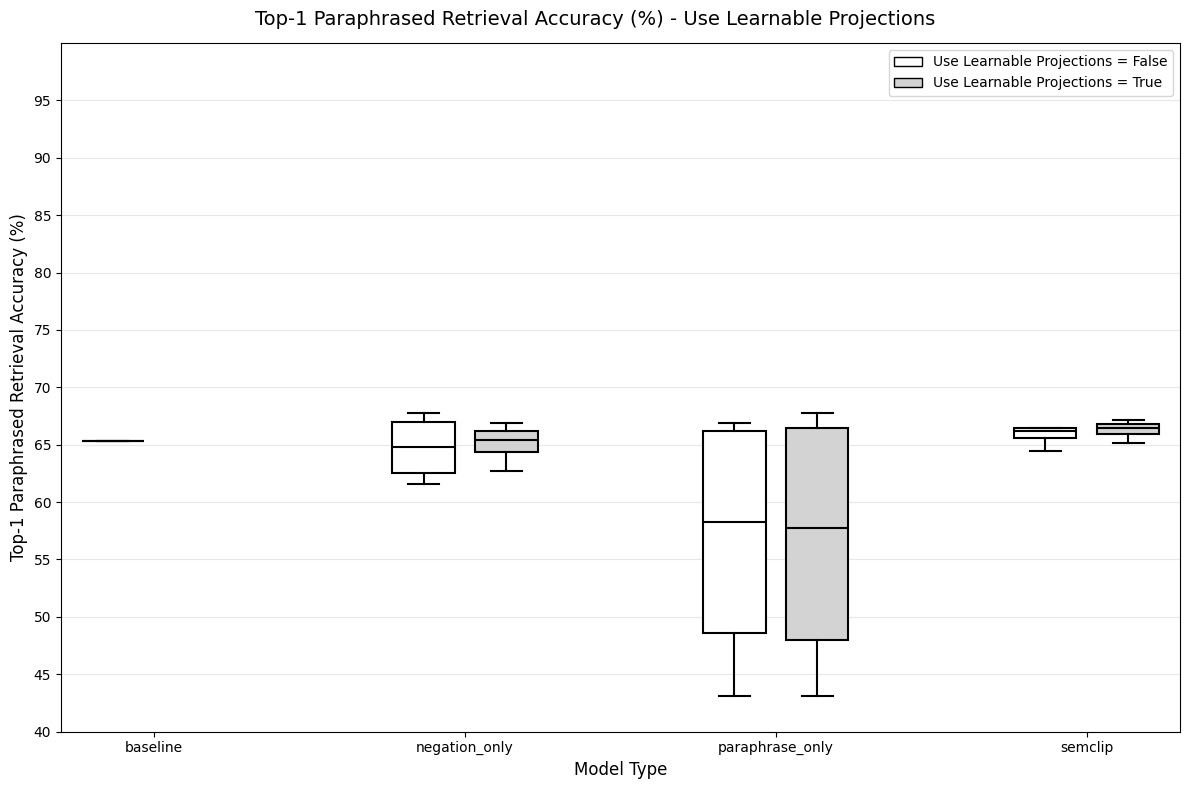}
    \par\vspace{5pt}
    \small (b) Effect of setting the use of learnable projection vectors on Top-1 accuracy using paraphrased caption for image matching.
    \label{fig:hp_ablation_learn_proj_para_scpp}
  \end{minipage}\par\vspace{10pt}

  \noindent
  \begin{minipage}{\linewidth}
    \centering
    \includegraphics[width=\linewidth,height=.23\textheight,keepaspectratio]{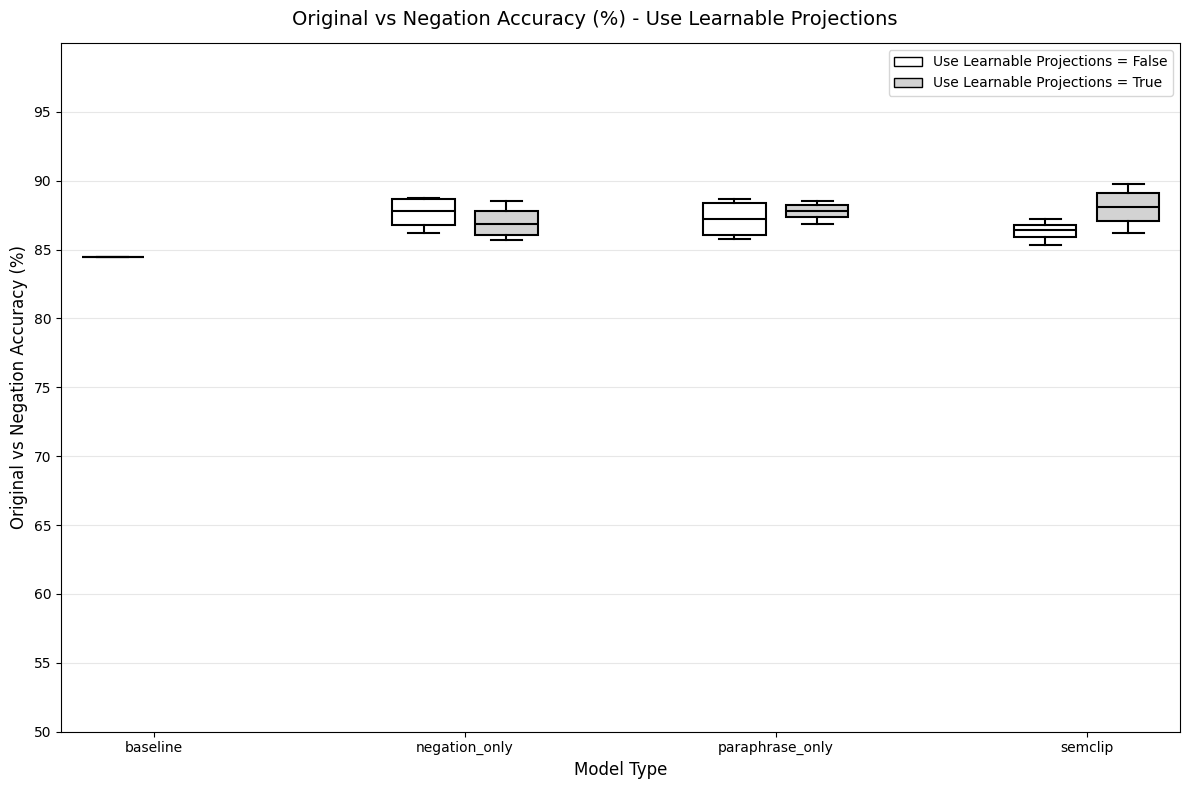}
    \par\vspace{5pt}
    \small (c) Effect of setting the use of learnable projection vectors on Top-1 accuracy using original caption over negated caption for image matching.
    \label{fig:hp_ablation_learn_proj_oon_scpp}
  \end{minipage}

  \caption{Effect of setting the use of learnable projection vectors on image matching accuracies using trained model (finetuned with Sugarcrepe\texttt{++} dataset \citep{Dumpala2024-hx}).}
  \label{fig:hyperparameters_ablation_learn_proj_scpp}
\end{figure}

\begin{figure}[htbp!]
  \centering
  \noindent
  \begin{minipage}{\linewidth}
    \centering
    \includegraphics[width=\linewidth,height=.23\textheight,keepaspectratio]{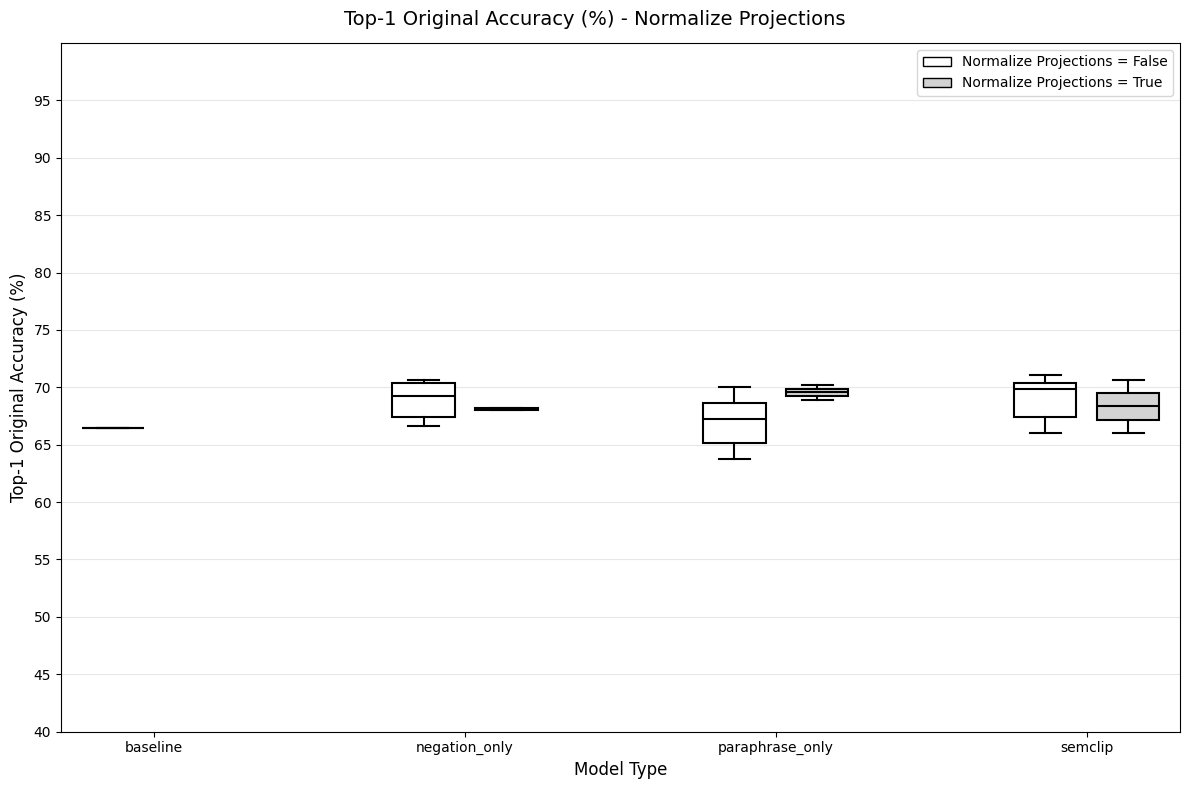}
    \par\vspace{5pt}
    \small (a) Effect of setting the projection vectors normalization on Top-1 accuracy using original caption for image matching.
    \label{fig:hp_ablation_norm_proj_orig_scpp}
  \end{minipage}\par\vspace{10pt}

  \noindent
  \begin{minipage}{\linewidth}
    \centering
    \includegraphics[width=\linewidth,height=.23\textheight,keepaspectratio]{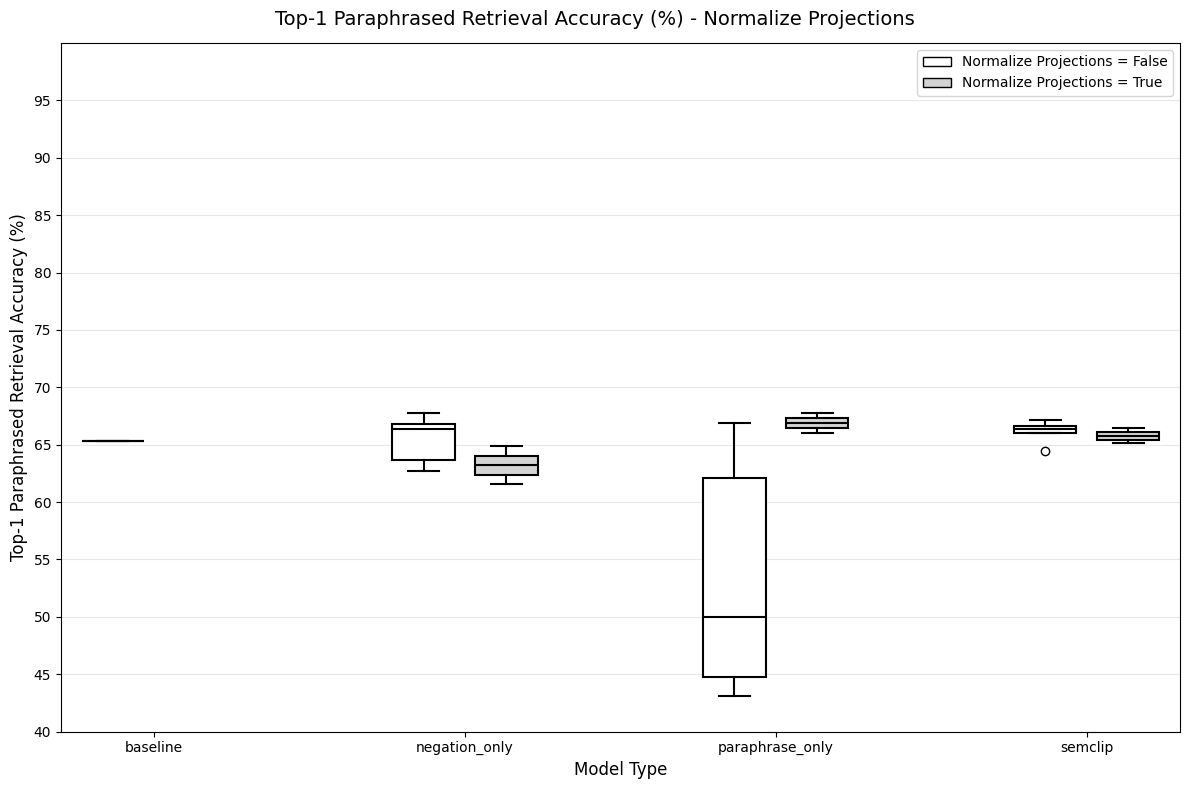}
    \par\vspace{5pt}
    \small (b) Effect of setting the projection vectors normalization on Top-1 accuracy using paraphrased caption for image matching.
    \label{fig:hp_ablation_norm_proj_para_scpp}
  \end{minipage}\par\vspace{10pt}

  \noindent
  \begin{minipage}{\linewidth}
    \centering
    \includegraphics[width=\linewidth,height=.23\textheight,keepaspectratio]{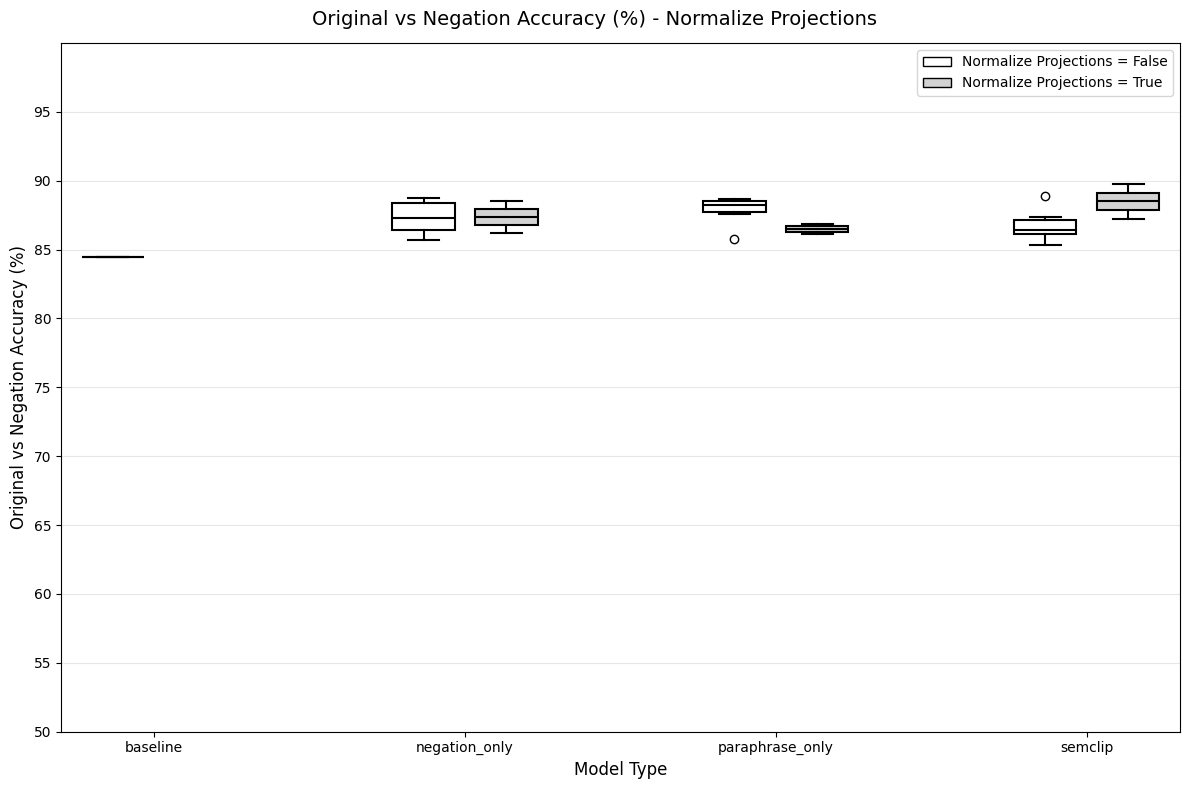}
    \par\vspace{5pt}
    \small (c) Effect of setting the projection vectors normalization on Top-1 accuracy using original caption over negated caption for image matching.
    \label{fig:hp_ablation_norm_proj_oon_scpp}
  \end{minipage}

  \caption{Effect of setting the projection vectors normalization on image matching accuracies using trained model (finetuned with Sugarcrepe\texttt{++} dataset \citep{Dumpala2024-hx}).}
  \label{fig:hyperparameters_ablation_norm_proj_scpp}
\end{figure}

\begin{figure}[htbp!]
  \centering
  \noindent
  \begin{minipage}{\linewidth}
    \centering
    \includegraphics[width=\linewidth,height=.23\textheight,keepaspectratio]{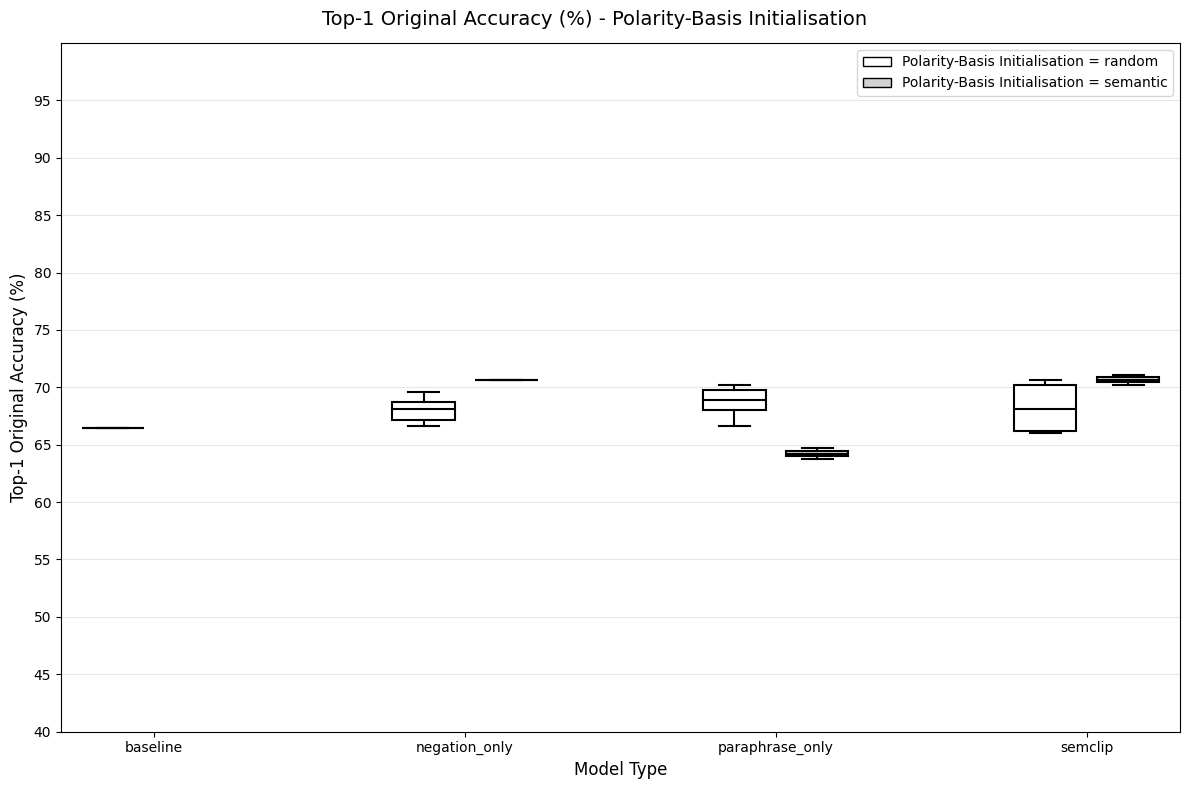}
    \par\vspace{5pt}
    \small (a) Effect of setting the polarity-basis initialisation on Top-1 accuracy using original caption for image matching.
    \label{fig:hp_ablation_basis_init_orig_scpp}
  \end{minipage}\par\vspace{10pt}

  \noindent
  \begin{minipage}{\linewidth}
    \centering
    \includegraphics[width=\linewidth,height=.23\textheight,keepaspectratio]{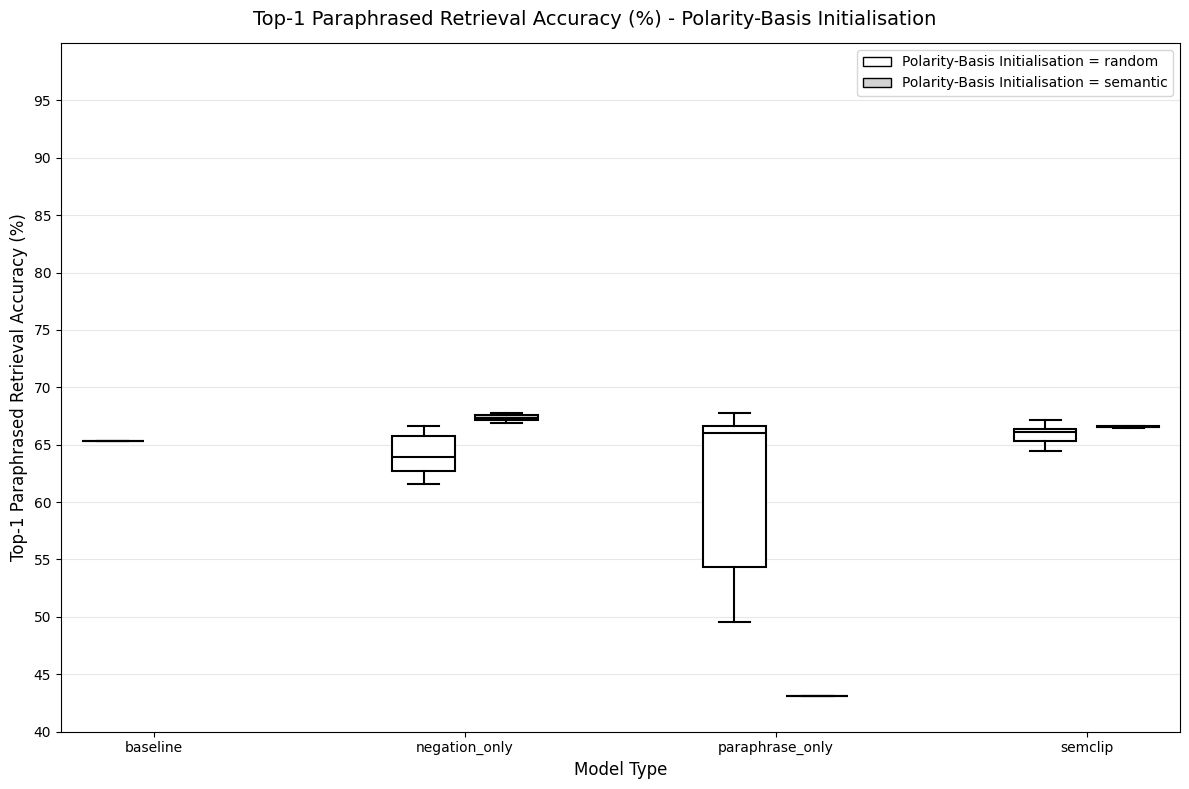}
    \par\vspace{5pt}
    \small (b) Effect of setting the polarity-basis initialisation on Top-1 accuracy using paraphrased caption for image matching.
    \label{fig:hp_ablation_basis_init_para_scpp}
  \end{minipage}\par\vspace{10pt}

  \noindent
  \begin{minipage}{\linewidth}
    \centering
    \includegraphics[width=\linewidth,height=.23\textheight,keepaspectratio]{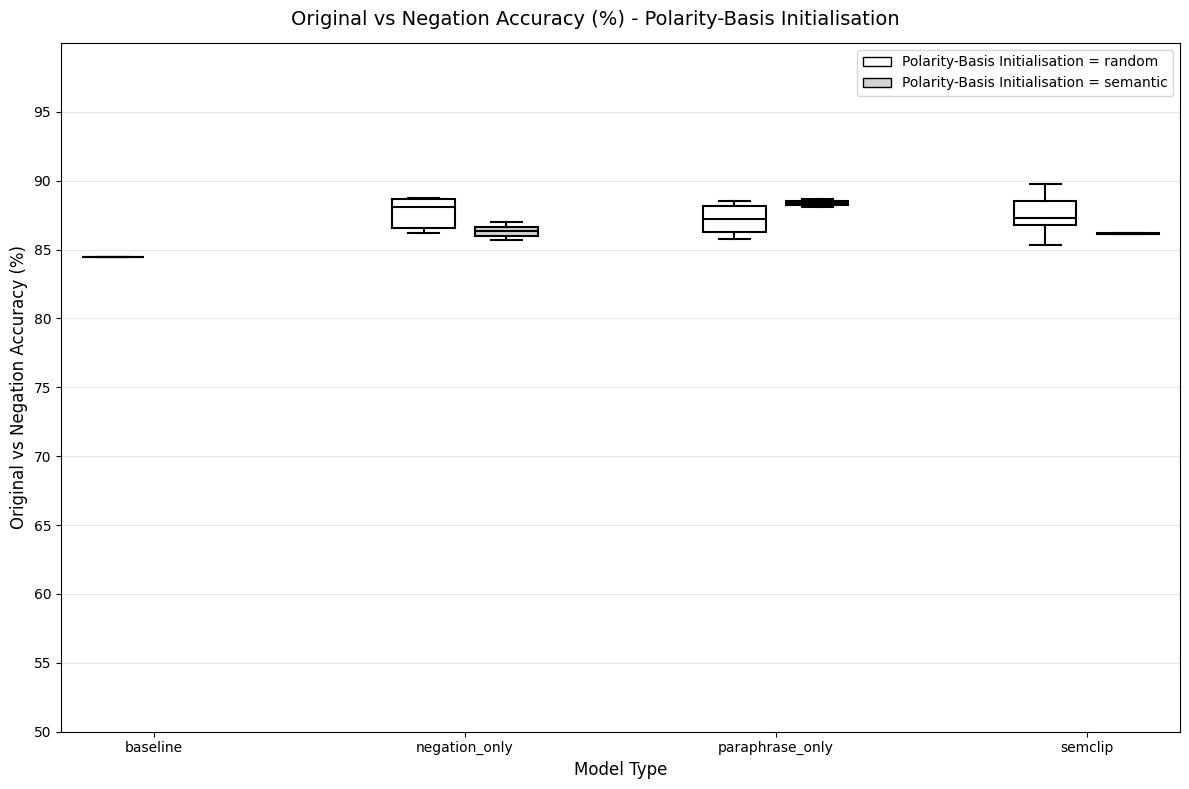}
    \par\vspace{5pt}
    \small (c) Effect of setting the polarity-basis initialisation on Top-1 accuracy using original caption over negated caption for image matching.
    \label{fig:hp_ablation_basis_init_oon_scpp}
  \end{minipage}

  \caption{Effect of setting the polarity-basis initialisation on image matching accuracies using trained model (finetuned with Sugarcrepe\texttt{++} dataset \citep{Dumpala2024-hx}).}
  \label{fig:hyperparameters_ablation_basis_init_scpp}
\end{figure}

Repeating the experiment with models fine-tuned on the Sugarcrepe\texttt{++} \citep{Dumpala2024-hx} dataset produces the results shown in  Figures~\ref{fig:hyperparameters_ablation_num_vec_scpp}, \ref{fig:hyperparameters_ablation_learn_proj_scpp}, \ref{fig:hyperparameters_ablation_norm_proj_scpp} and \ref{fig:hyperparameters_ablation_basis_init_scpp}. SemCLIP shows consistent performance across all image matching tasks. On original captions, SemCLIP achieves a range of 66.0\% to 71.1\% accuracy, while maintaining robustness on paraphrased captions (with performance near to that of using original captions at a range of 64.4\% and 67.1\%). On the negation task, all models perform well (84.4 - 89.8\%), with SemCLIP offering the peak performance of 89.8\%.

Overall, these findings highlight SemCLIP's ability to learn robust representations to paraphrasing and negation without much performance degradation on the original task. Among the hyperparameters studied, the number of projection vectors is worth further investigation, offering a potential direction for future fine-tuning and performance optimization.

\subsection{Analysis of standard accuracy and negated accuracy diagrams on downstream classification tasks} 

\begin{figure}[htbp!]
  \centering
  \noindent
  \begin{minipage}{\linewidth}
    \centering
    \includegraphics[width=\linewidth,height=.26\textheight,keepaspectratio]{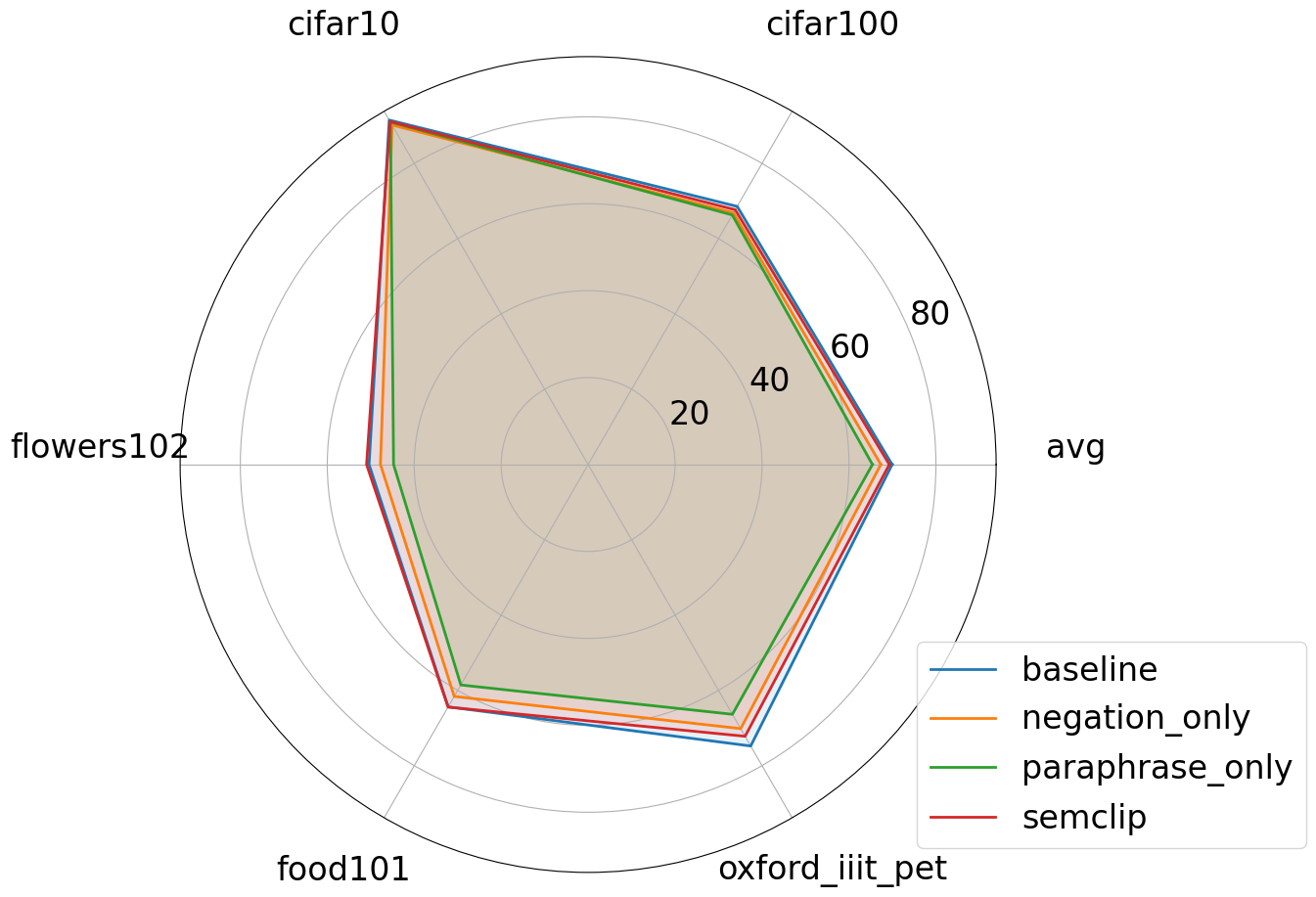}
    \par\vspace{5pt}
    \small (a) Mean Standard Accuracy on different classification tasks. Higher is better.
    \label{fig:class_std_acc_ccneg}
  \end{minipage}\par\vspace{10pt}

  \noindent
  \begin{minipage}{\linewidth}
    \centering
    \includegraphics[width=\linewidth,height=.26\textheight,keepaspectratio]{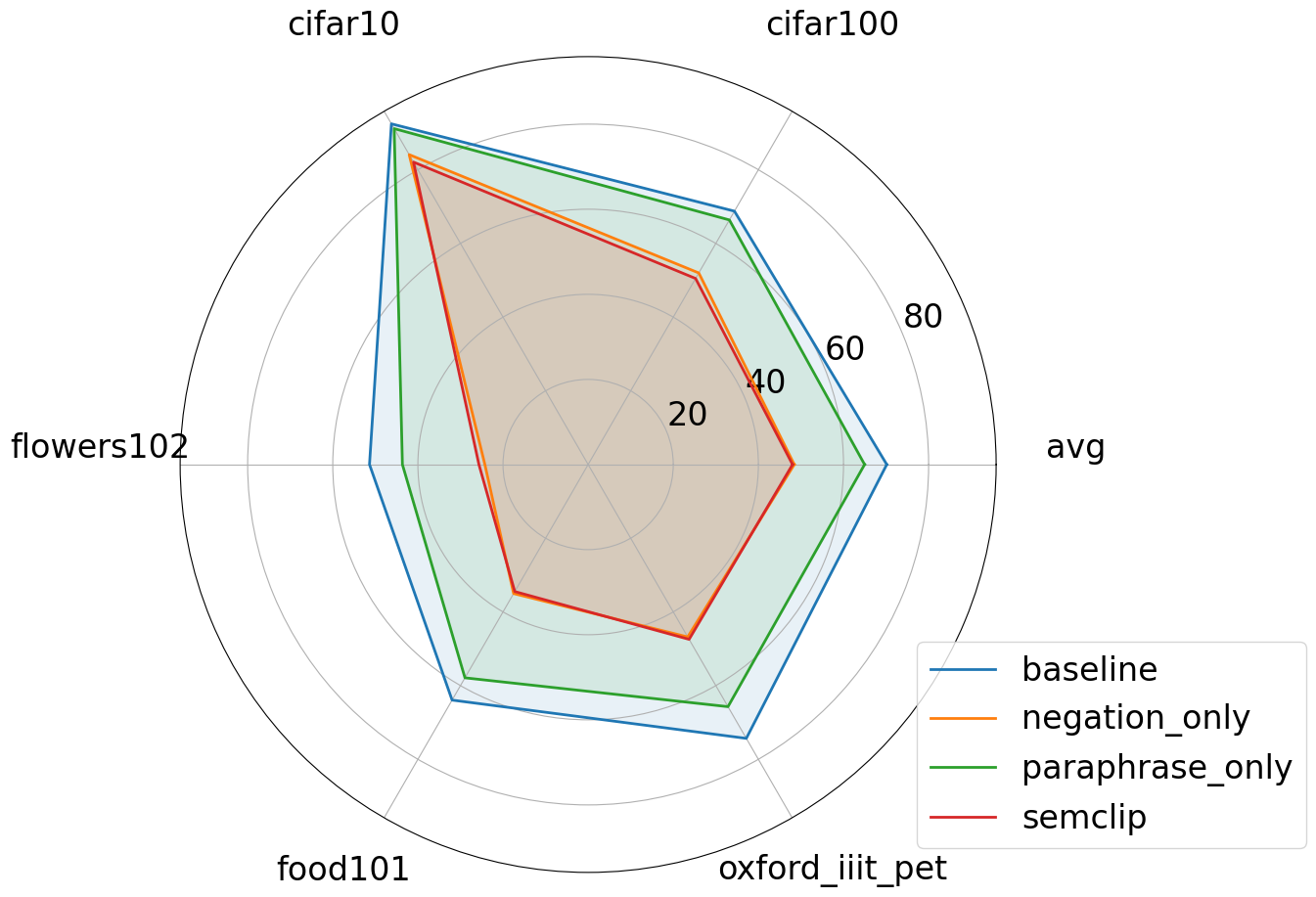}
    \par\vspace{5pt}
    \small (b) Mean Negated Accuracy on different classification tasks. Lower is better.
    \label{fig:class_neg_acc_ccneg}
  \end{minipage}\par\vspace{10pt}

  \noindent
  \begin{minipage}{\linewidth}
    \centering
    \includegraphics[width=\linewidth,height=.26\textheight,keepaspectratio]{images/nesy_classification_delta_ccneg.png}
    \par\vspace{5pt}
    \small (c) Mean Accuracy Delta on different classification tasks. Higher indicates better differentiation.
    \label{fig:class_delta_ccneg}
  \end{minipage}

  \caption{Mean accuracy on downstream classification tasks by models with different training loss terms (finetuned with CCNeg dataset \citep{singh2024learn}).}
  \label{fig:downstream_class_acc_ccneg}
\end{figure}
\begin{figure}[htbp!]
  \centering
  \noindent
  \begin{minipage}{\linewidth}
    \centering
    \includegraphics[width=\linewidth,height=.26\textheight,keepaspectratio]{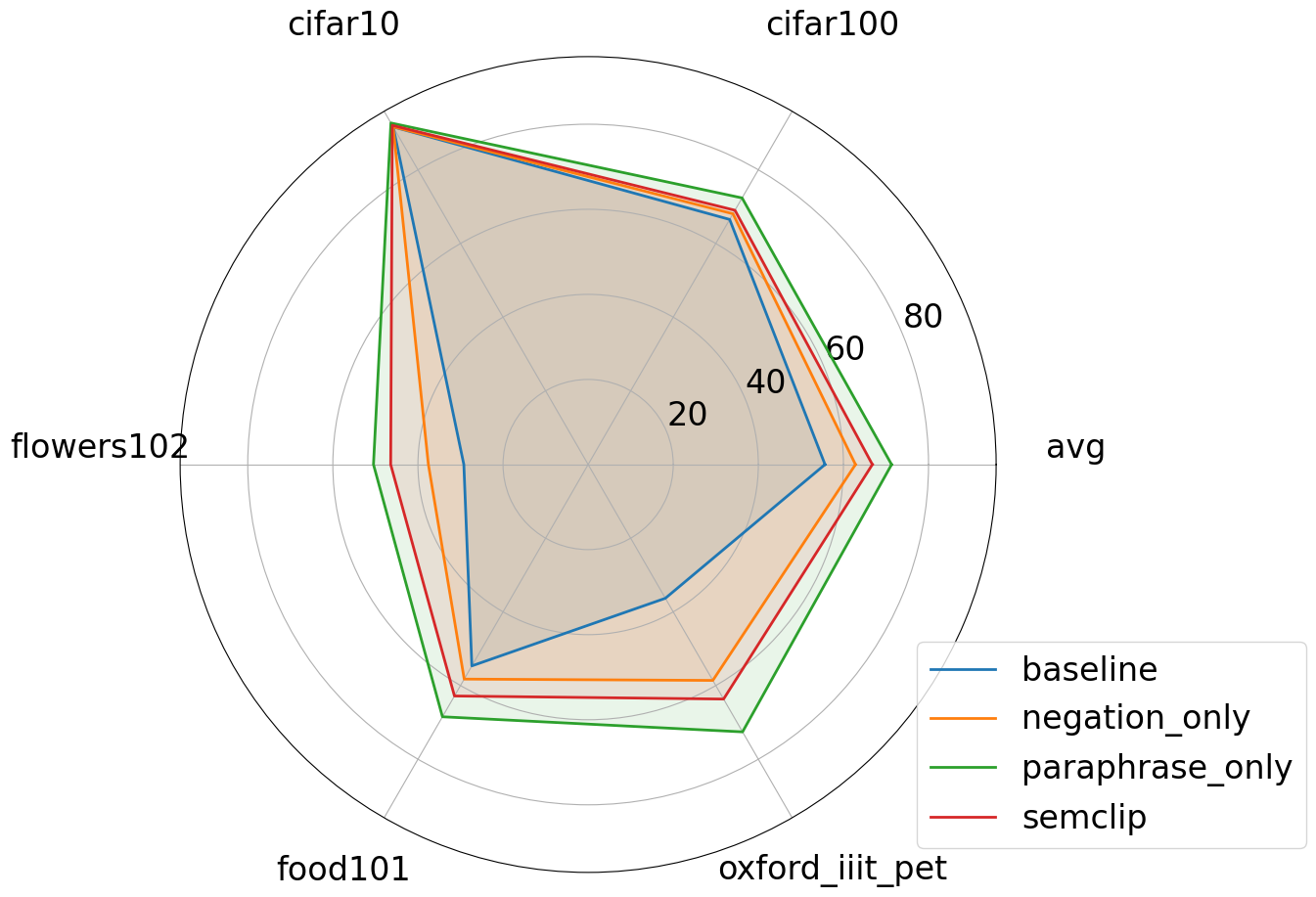}
    \par\vspace{5pt}
    \small (a) Mean Standard Accuracy on different classification tasks. Higher is better.
    \label{fig:class_std_acc_scpp}
  \end{minipage}\par\vspace{10pt}

  \noindent
  \begin{minipage}{\linewidth}
    \centering
    \includegraphics[width=\linewidth,height=.26\textheight,keepaspectratio]{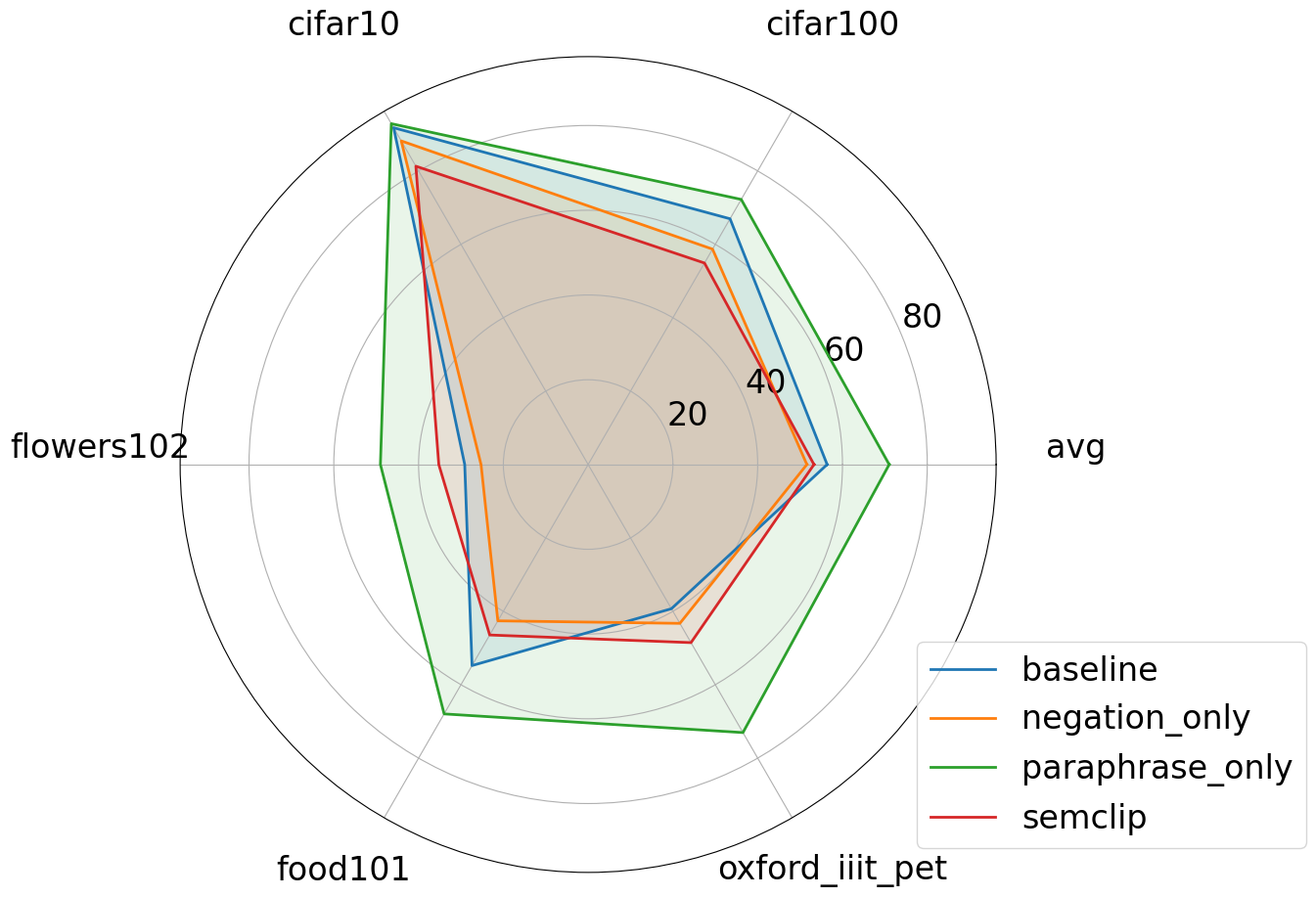}
    \par\vspace{5pt}
    \small (b) Mean Negated Accuracy on different classification tasks. Lower is better.
    \label{fig:class_neg_acc_scpp}
  \end{minipage}\par\vspace{10pt}

  \noindent
  \begin{minipage}{\linewidth}
    \centering
    \includegraphics[width=\linewidth,height=.26\textheight,keepaspectratio]{images/nesy_classification_delta_scpp.png}
    \par\vspace{5pt}
    \small (c) Mean Accuracy Delta on different classification tasks. Higher indicates better differentiation.
    \label{fig:class_delta_scpp}
  \end{minipage}

  \caption{Mean accuracy on downstream classification tasks by models with different training loss terms (finetuned with Sugarcrepe\texttt{++} dataset \citep{Dumpala2024-hx}).}
  \label{fig:downstream_class_acc_scpp}
\end{figure}

Figures~\ref{fig:downstream_class_acc_ccneg} and \ref{fig:downstream_class_acc_scpp} present the classification accuracy on (a) original caption and (b) negated caption, as well as (c) the accuracy delta between (a) and (b). SemCLIP has shown an improved robustness to negation overall as it has the largest delta on four of five tasks when trained with both CC-Neg and SCPP dataset.

SemCLIP fine-tuned on the CC-Neg dataset obtained the following image classification accuracies on downstream tasks: CIFAR10 91.1\%, CIFAR100 67.7\%, Flower102 51.0\%, Food101 64.4\%, and Oxford-IIIT Pet 72.2\%. Fine-tuned on the Sugarcrepe\texttt{++} dataset, SemCLIP's classification accuracies were: CIFAR10 92.1\%, CIFAR100 69.0\%, Flower102 46.4\%, Food101 62.8\%, and Oxford-IIIT Pet 63.7\%. The variations of performance observed here - very small variation for CIFAR10 and considerable variation in the case of Oxford-IIIT Pet — may depend of course on the fine-tuning but also on the CLIP pre-training. These downstream tasks were the same as investigated by CoN-CLIP, c.f. \citep{singh2024learn}, which used CC-Neg only for fine-tuning. Direct comparison between SemCLIP and CoN-CLIP were not carried out. Future comparative evaluations could consider fine-tuning CoN-CLIP with Sugarcrepe\texttt{++}.



\stop
\end{document}